\documentclass[10pt,twocolumn,letterpaper]{article}

\usepackage{cvpr}
\usepackage{times}
\usepackage{epsfig}
\usepackage{graphicx}
\usepackage{amsmath}
\usepackage{amssymb}
\usepackage{epstopdf}
\usepackage[FIGTOPCAP]{subfigure}
\usepackage[percent]{overpic}
\usepackage{xcolor}
\usepackage{pict2e}
\usepackage{algorithm,algorithmic}

\usepackage[breaklinks=true,bookmarks=false]{hyperref}

\cvprfinalcopy 

\def\cvprPaperID{5176} 

\ifcvprfinal\fi

\begin{document}

\title{Side Window Filtering}

\author{Hui Yin$^{a}$\thanks{authors equally contribute to this paper}\quad Yuanhao Gong$^{a*}$\quad Guoping Qiu$^{a,b}$\\
$^a$ College of Information Engineering and Guangdong Key Lab for Intelligent Information Processing,\\
Shenzhen University, China $^b$ School of Computer Science, The University of Nottingham, UK\\
{\tt\small yinhui0606@gmail.com}
{\tt\small gong@szu.edu.cn}
{\tt\small guoping.qiu@nottingham.ac.uk}
}

\maketitle
\thispagestyle{empty}

\begin{abstract}
   Local windows are routinely used in computer vision and almost without exception the center of the window is aligned with the pixels being processed. We show that this conventional wisdom is not universally applicable. When a pixel is on an edge, placing the center of the window on the pixel is one of the fundamental reasons that cause many filtering algorithms to blur the edges. Based on this insight, we propose a new Side Window Filtering (SWF) technique which aligns the window's side or corner with the pixel being processed. The SWF technique is surprisingly simple yet theoretically rooted and very effective in practice. We show that many traditional linear and nonlinear filters can be easily implemented under the SWF framework. Extensive analysis and experiments show that implementing the SWF principle can significantly improve their edge preserving capabilities and achieve state of the art performances in applications such as image smoothing, denoising, enhancement, structure-preserving texture-removing, mutual-structure extraction, and HDR tone mapping. In addition to image filtering, we further show that the SWF principle can be extended to other applications involving the use of a local window. Using colorization by optimization as an example, we demonstrate that implementing the SWF principle can effectively prevent artifacts such as color leakage associated with the conventional implementation. Given the ubiquity of window based operations in computer vision, the new SWF technique is likely to benefit many more applications.
\end{abstract}

\section{Introduction}
\begin{figure}[!tbh]
	\centering
	\subfigure[Step Edge]{\includegraphics[height=0.98in]{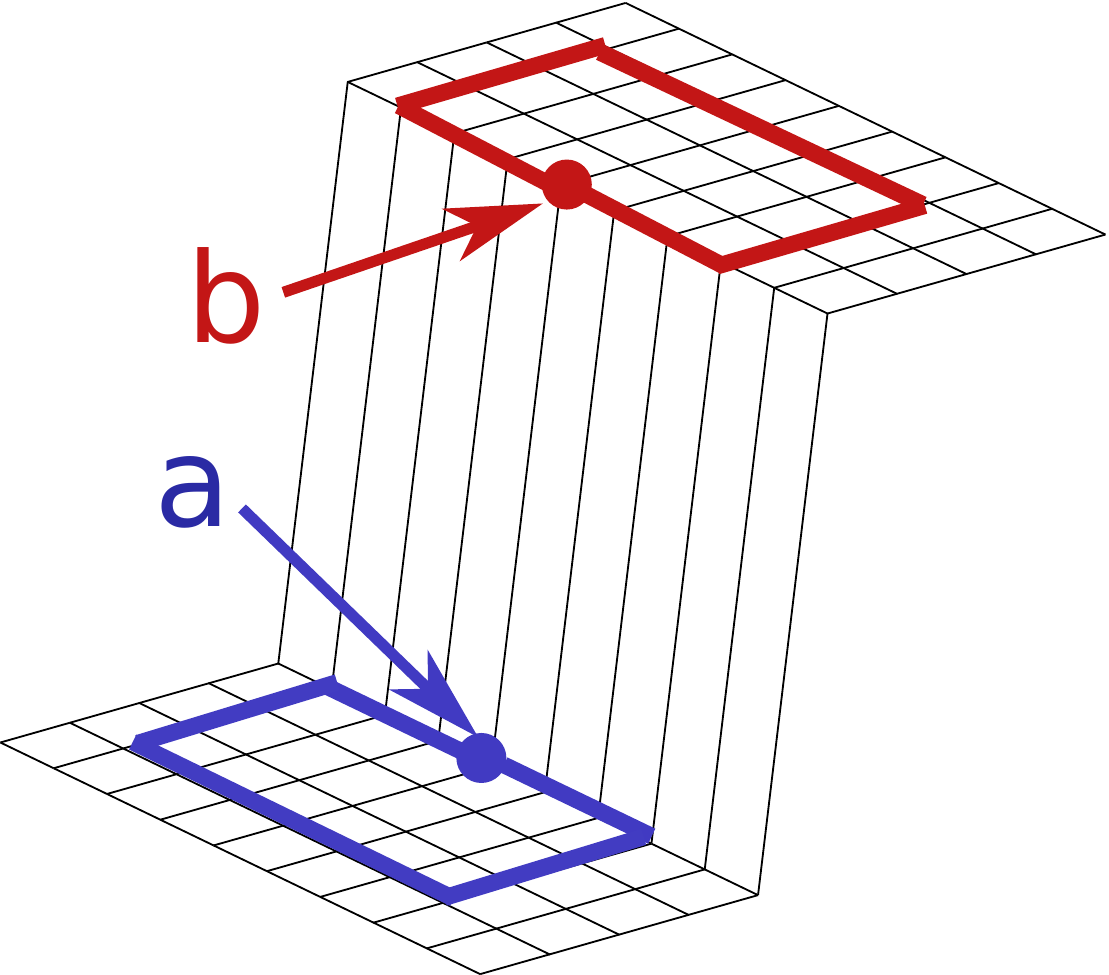}}
	\subfigure[Ramp Edge]{\includegraphics[height=0.98in]{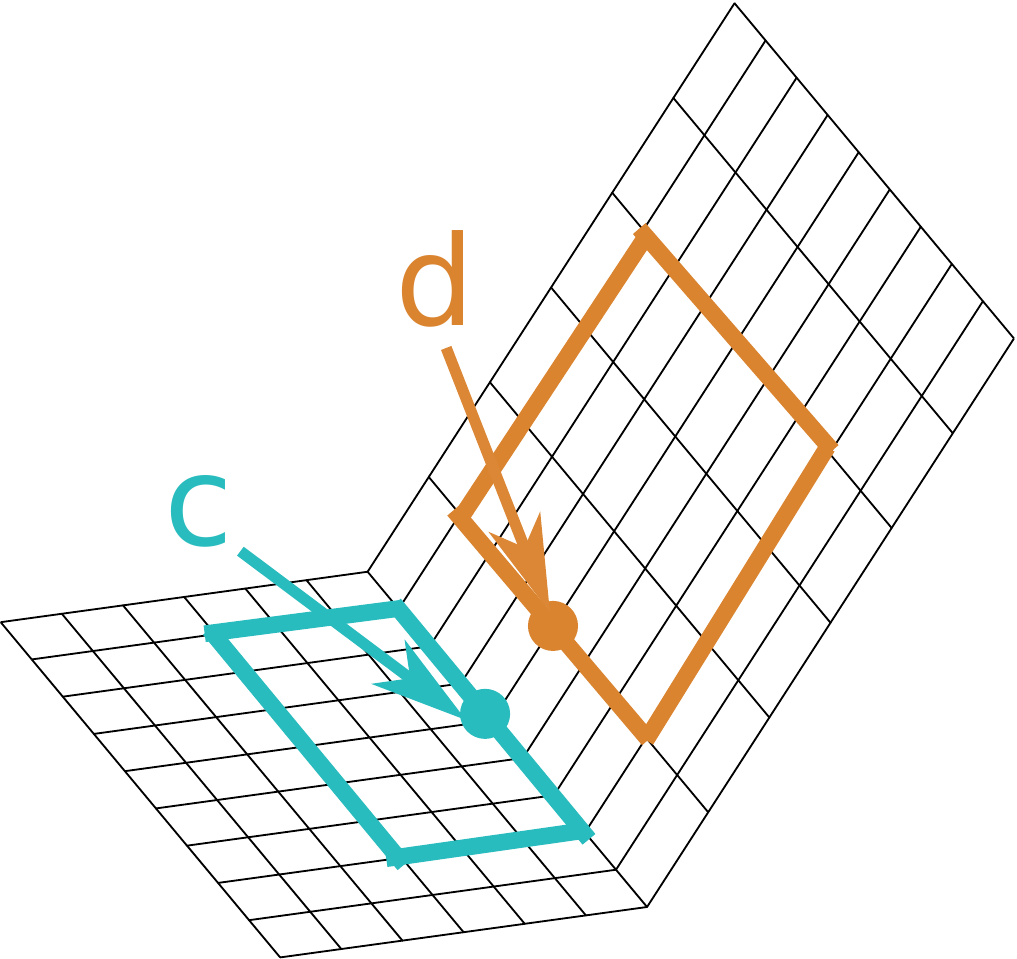}}
	\subfigure[Roof Edge]{\includegraphics[height=0.98in]{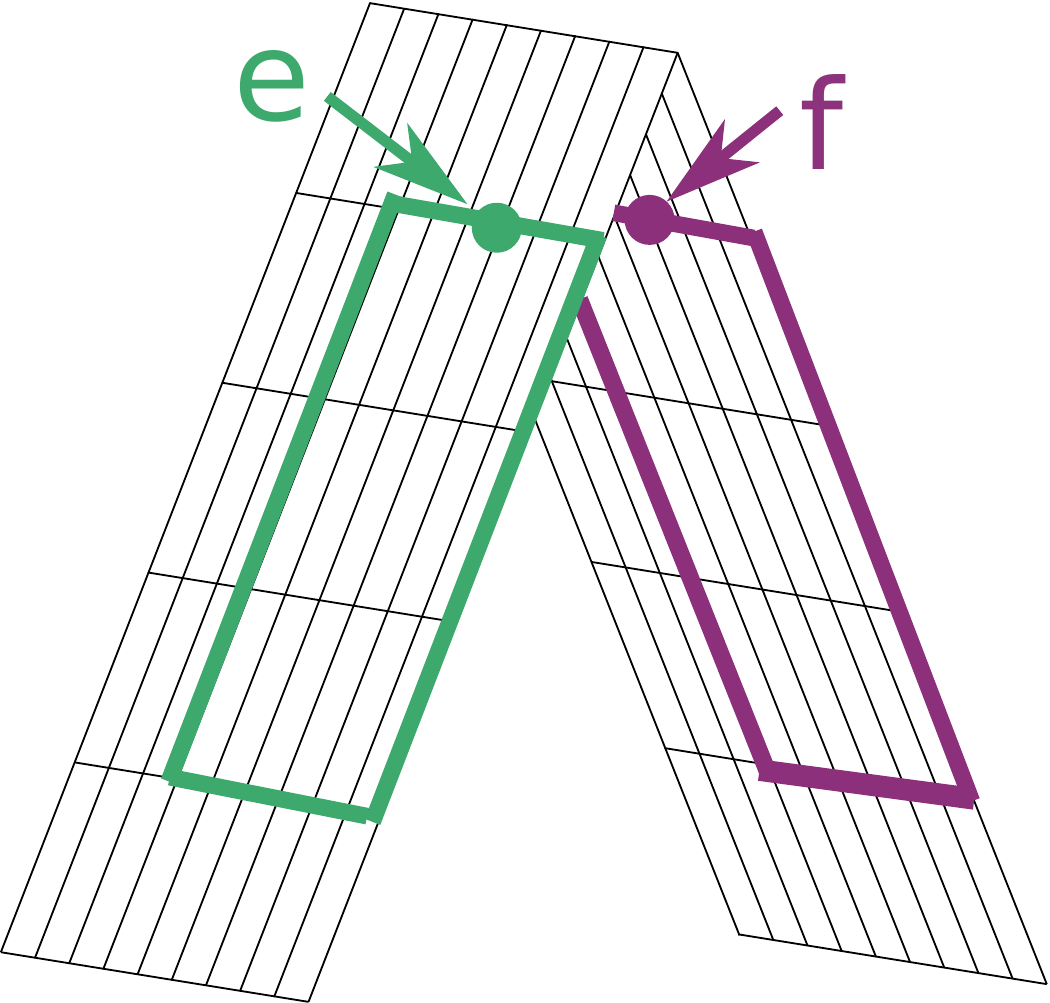}}
	\caption{Model of ideal edges in 2D piecewise images. The pixel `a'$\sim$`f' are on edges or near edges. To satisfy the linear assumption, they should be approximated in the side windows which have the same colors with them, not the local windows centered at them.}
	\label{edgeanalysis1}
\end{figure}
In the fields of computational photography and image processing, many applications involve the concept of image filtering to denoise \cite{denoise}, deblur \cite{deblur} and enhance details \cite{details}. For decades, various filters have been developed, such as box filter, Gaussian filter and median filter, to name a few. These filters are widely used in image deblurring and sharpening, edge detection and feature extraction \cite{book1}.

There are many applications require image filtering that can preserve edges. Typical examples include tone mapping of high dynamic range (HDR) images \cite{hdr}, detail enhancement via multi-lighting images \cite{enhancement}, and structure-preserving and texture removing \cite{rtv}\cite{rgf}.

For this reason, many edge-preserving filters have been proposed. Basically, these edge-preserving filters can be divided into two categories. One is global optimization based algorithms, such as the total variation (TV) algorithm \cite{denoise}, its iterative shrinkage approach \cite{tv1}, the relative total variation  algorithm \cite{rtv} and the weighted least squares algorithm \cite{wls}. The other is local optimization based algorithms, such as bilateral filter \cite{bf}, its accelerated versions \cite{fbf}\cite{fbf1}\cite{fbf2}, guided filter \cite{gf}, its extensions \cite{wgf}\cite{ggf}, rolling guidance filter \cite{rgf}, mutual structure joint filtering \cite{msfjf} and curvature filter \cite{details}. In general, the local based filters can be calculated in real time. This is preferred because many real application scenarios require real-time processing. 

\subsection{Filtering Fundamentals}
Local based filters always attempt to estimate an output of a pixel based on its neighbors. Almost without exception, the pixel being processed is located at the center of an operation window and other pixels in the operation window are its neighbors. Basically, there are two ways to do estimation: linear approximation, such as box filter and Gaussian filter, and non-linear approximation, such as median filter \cite{mf}, bilateral filter \cite{bf} and guided filter \cite{gf}.  

A common linear approximation based image filtering operation assumes that the image is piecewise linear and approximate a pixel as the weighted average of its neighbor pixels over a local window 
\begin{equation}
I_i^{'}=\sum_{j\in \Omega_i}{\omega_{ij}q_j}
\end{equation}
where $\Omega_i$ is the local window (support region) centered at the pixel $i$, $\omega_{ij}$ denotes the weight kernel, 
$q_i$ and $I_i$ are the intensities of the input image $q$ and the output image $I$ at location $i$, respectively. 

The discrepancy between the filter output and the original image can be formulated as the following cost function
\begin{equation}
E_i = ||I_i - I_i^{'}||_2^2=(I_i - \sum_{j\in \Omega_i}{\omega_{ij}q_j})^2
\end{equation}
Different weight kernels will result in different filtering output images and in most cases the task of designing a filtering algorithm is that of estimating the weights. 
Often there is a trade off between manipulating the input image towards a desired target and keeping it close to the original. It is worth noting that optimization problem of the form similar to eq. (2) is found in many applications including colorization \cite{colorization}\cite{colorizationofqiu} and image segmentation \cite{seg1}\cite{seg2}, where the weight functions are usually referred to as affinity functions. Nonlinear approximation filtering such as median filtering can also be formulated as a similar form of optimization problem \cite{med1}. 

\subsection{Problem and Motivation}
In many applications that using the form of filtering algorithm in eq. (1), it is desired to smooth out genuine noise and at the same time preserve edges and other signal details.  
For analysis convenience, we focus our study on three types of typical edges \cite{cie}, step edge, ramp edge and roof edge, and model them in 2D signal space as shown in Fig.~\ref{edgeanalysis1}. We use $g(x,y)$ to denote the intensity value at $(x,y)$. The functions $g(x,y)$ shown in this figure are continuous but non-differentiable. Considering the locations where the intensity changes (an edge), for example, at location `a'. We use `a-' and `a+' to denote the left limit ($x-\epsilon,y$) and right limit ($x+\epsilon,y$), respectively, where $\epsilon>0$. Clearly, $g(x-\epsilon,y)\neq g(x+\epsilon,y)$ and (or) $g'(x-\epsilon,y)\neq g'(x+\epsilon,y)$ due to the edge jump. Therefore, the Taylor expansion at these two regions are different: $g(x-2\epsilon,y)\approx g(x-\epsilon,y)+g'(x-\epsilon,y)(-\epsilon)$ and $g(x+2\epsilon,y)\approx g(x+\epsilon,y)+g'(x+\epsilon,y)\epsilon$. Therefore, any approximation at location `a-' must come from the left regions of `a' while any approximation at location `a+' must come from the right regions of `a'. Similar statements apply for other edge locations such as `b', `c', and `d' in Fig.~\ref{edgeanalysis1}.

Based on the analysis and eq. (1), if a pixel $i$ is on an edge, the support region $\Omega_i$ must be restricted to one side of the edge, otherwise, it is not possible to use a linear combination of the neighbors to approximate $i$. In other words, we cannot place the center of $\Omega_i$ over $i$ but rather we must place the side of $\Omega_i$ over $i$. Inspired by this discovery, a new edge-preserving strategy, termed side window filtering (SWF) technique, is proposed. We consider each target pixel as a potential edge and generate multiple local windows (named as side windows) around it, each of which aligns the target pixel with a side or a corner (instead of the center) of the window. The output of SWF is a linear combination of the neighbors in one of the side windows which can best approximate the target pixel. 

\subsection{Our Contributions}
The novel contributions of this paper are:

1. Using Taylor expansion, we show that in order to reconstruct an edge pixel using a linear combination of its neighbors, the neighbor pixels must come from one side of the edge. Based on this insight, we propose the side window filtering (SWF) technique as an effective and practical edge preserving filtering solution. 

2. We show how traditional linear filters such as box filter and Gaussian filter, popular non-linear filters such as median filter, bilateral filter and  guided filter can easily be implemented under the SWF framework. Through extensive analysis, we show that implementing these popular filters based on the new SWF framework can significantly improve their edge preserving capabilities.

3. We show that the implementing traditional filters under the new SWF framework provides state of the art performances in a variety of real world applications including image smoothing, denoising, enhancement,  structure-preserving texture-removing, mutual-structure extraction, and high dynamic range image tone mapping.

4. We show that the new SWF framework can be extended to other applications involving a local window and a linear combination of a neighborhood of pixels. Using colorization by optimization as an example we demonstrate that implementing the SWF principle can effectively prevent artifacts such as color leakage.

5. The SWF technique is very simple but theoretically rooted and in practice surprisingly effective. Given the ubiquity of window based operations, the SWF principle has the potential of benefiting many areas in image processing and computer vision.  


\begin{figure*}[htbp] 
	\centering
	\subfigure[]{\includegraphics[height=1.3in]{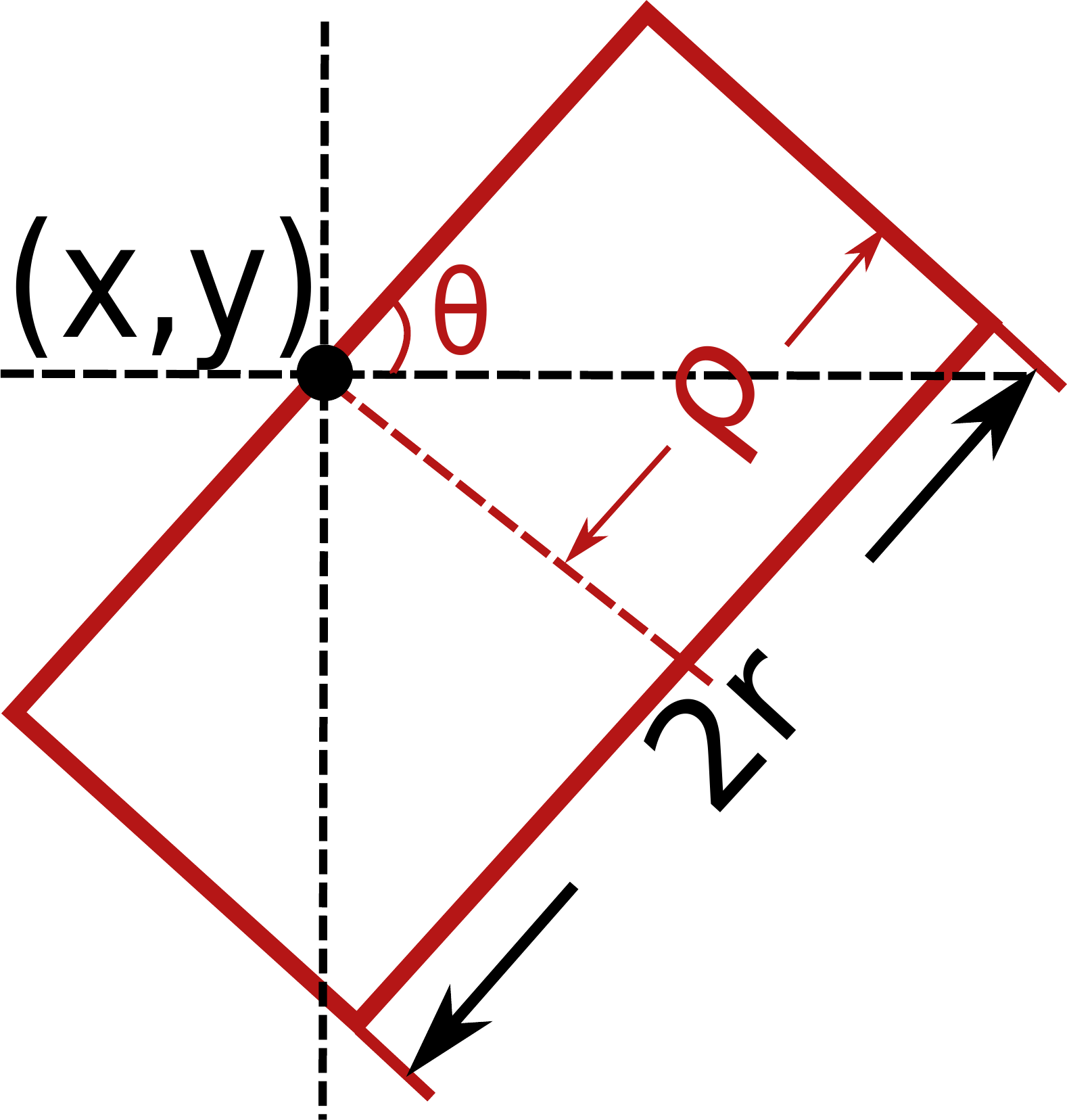}}
	\subfigure[]{\includegraphics[height=1.3in]{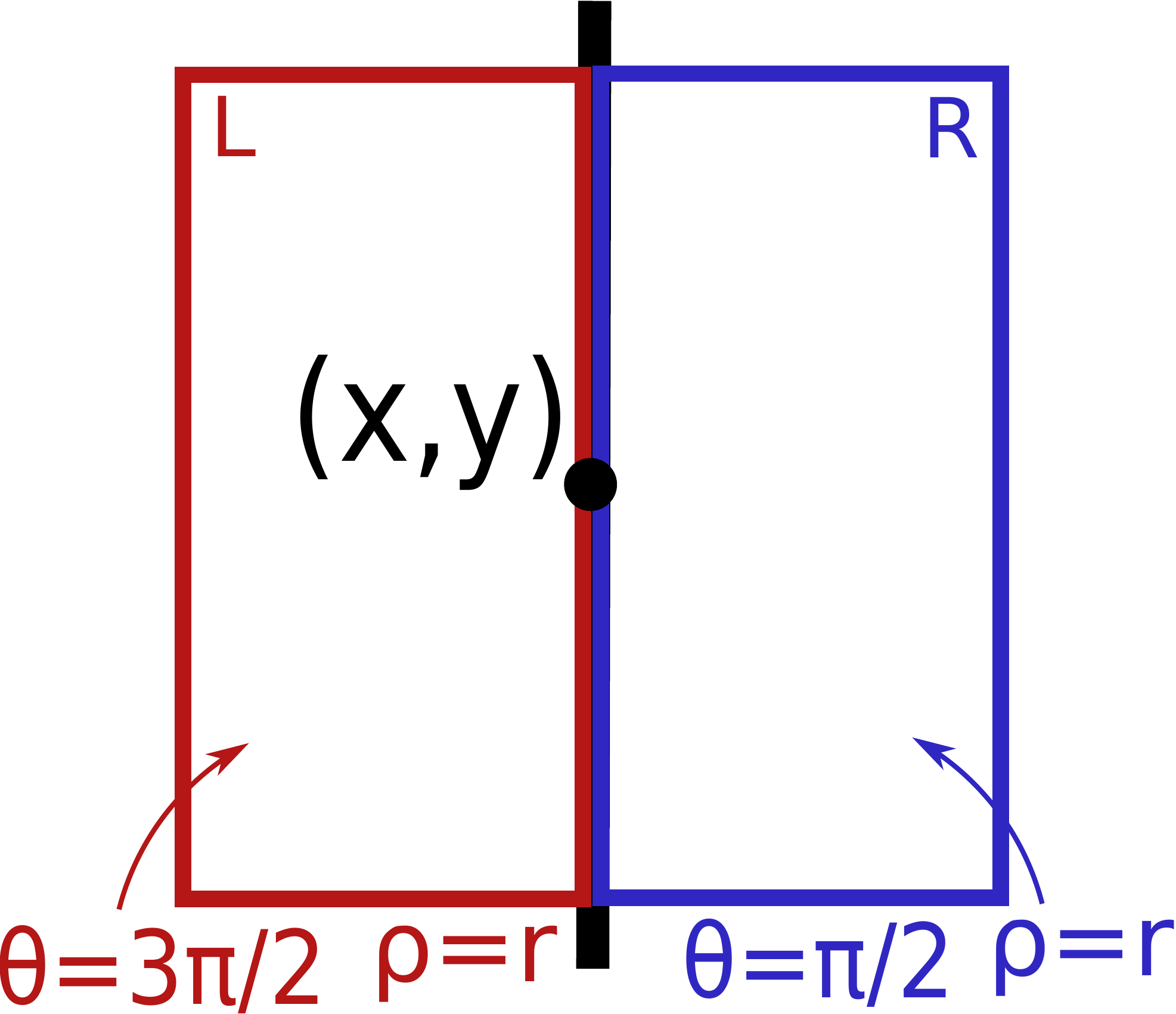}}
	\subfigure[]{\includegraphics[height=1.3in]{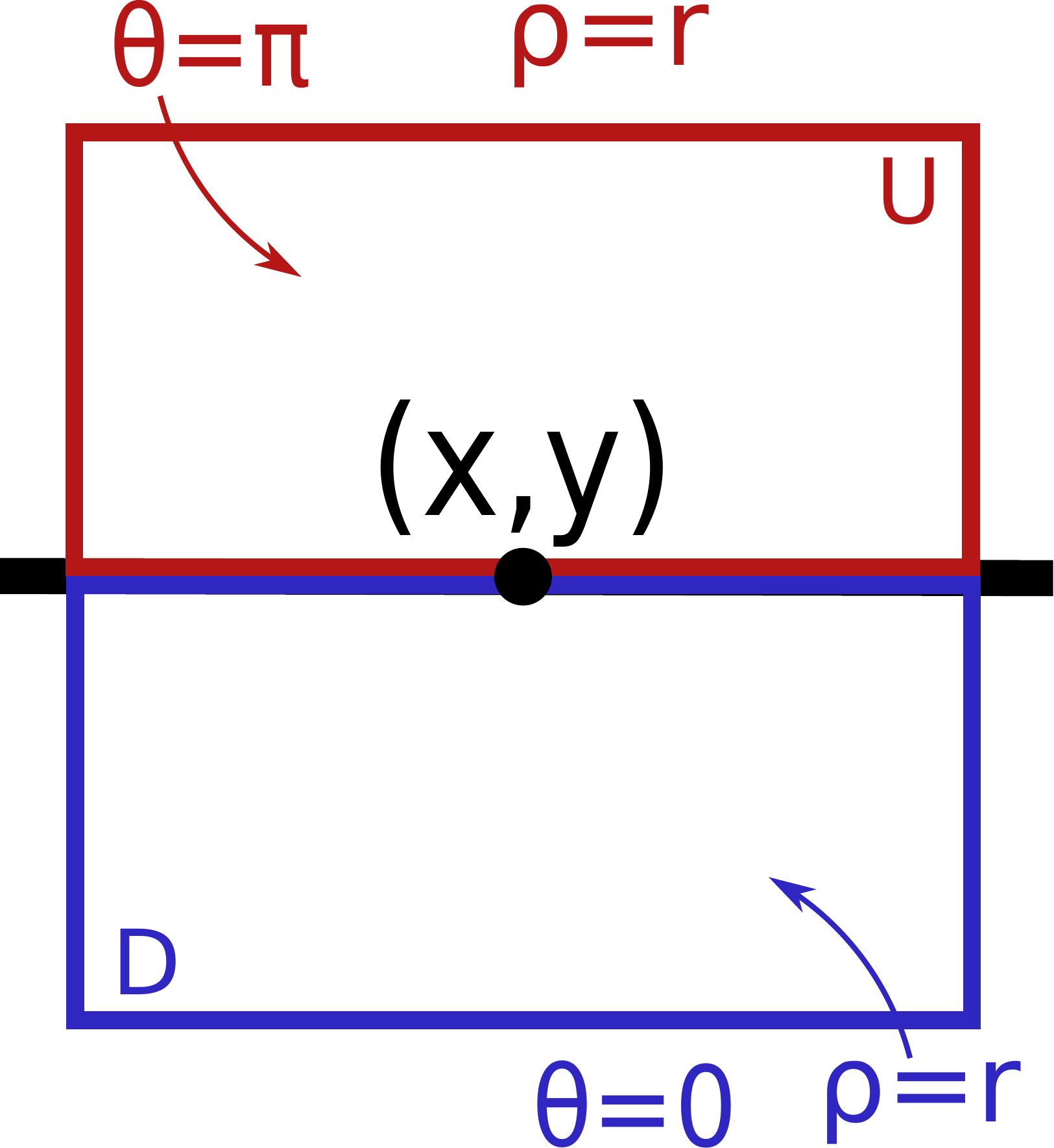}}
	\subfigure[]{\includegraphics[height=1.3in]{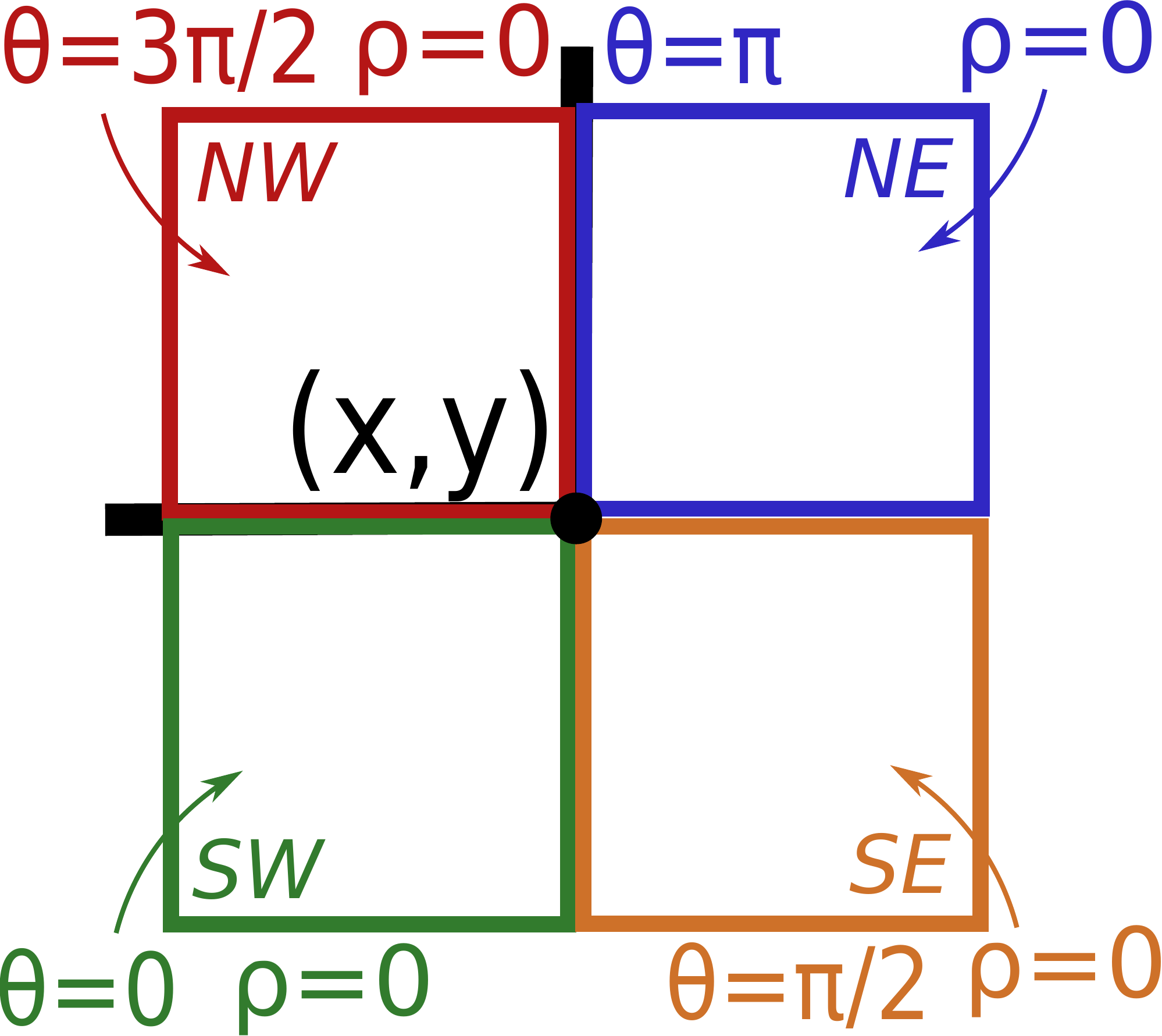}}
	\caption{Definition of side window. $r$ is the radius of the window. (a) The definition of side window in continuous case. (b) The $left$ (red rectangle) and $right$ (blue rectangle) side windows. (c) The $up$ (red rectangle) and $down$ (blue rectangle) side windows. (d) The $northwest$ (red rectangle), $northeast$ (blue rectangle), $southwest$ (green rectangle) and $southeast$ (orange rectangle) side windows.}
	\label{eightwindows}
\end{figure*}

\section{Side Window Filtering Technique}
First of all we define the side window in a continuous case. The definition of a side window is shown in Fig.~\ref{eightwindows}(a), with parameters $\theta$ and $r$. $\theta$ is the angle between the window and the horizontal line, $r$ is the radius of the window, $\rho\in\{0, r\}$ and $(x,y)$ is the position of the target pixel $i$. $r$ is a user-defined parameter and it will be fixed for all the side windows. By changing $\theta$ and fixing $(x,y)$, we can change the direction of the window while aligning its side with $i$. 

To simplify the calculation, we only define eight side windows in a discrete case, as shown in Fig.~\ref{eightwindows}(b)$\sim$(d). These eight specific windows correspond to $\theta=k\times\frac{\pi}{2}, k\in[0,3]$. By setting $\rho = r$, we have the $down (D)$ , $right (R)$, $up (U)$, $left (L)$ side windows, denoted as $\omega^{D}_i$, $\omega^{R}_i$, $\omega^{U}_i$ and $\omega^{L}_i$. They align $i$ with their sides. By setting $\rho = 0$, we have the southwest ($SW$), southeast ($SE$), northeast ($NE$) and northwest ($NW$) side windows, as shown in Fig.~\ref{eightwindows}(d) and denoted as $\omega^{SW}_i$, $\omega^{SE}_i$, $\omega^{NE}_i$ and $\omega^{NW}_i$. They align $i$ with their corners. It is worth pointing out that there is significant flexibility in designing the size, shape and orientation of the side windows. And the only specific requirement is that the pixel under consideration is placed on the side or corner of the window.
\begin{algorithm}[!htb]
	\caption{Calculate the SWF for each pixel} \label{algo_swf}
	\begin{algorithmic}[1]
		\REQUIRE $w_{ij}$ is the weight of pixel $j$, which is in the neighborhood of the target pixel $i$, based on kernel function $F$. $S = \{L, R, U, D, NW, NE, SW, SE\}$ is the set of side window index.
		\STATE $I_{\mathrm{{n}}} = \frac{1}{N_{\mathrm{{n}}}}\mathop{\sum}_{j\in\omega^{\mathrm{{n}}}_i}w_{ij}q_j$, $N_{\mathrm{{n}}} = \mathop{\sum}_{j\in\omega^{\mathrm{{n}}}_i}w_{ij}, \mathrm{{n}} \in S $
		\if false
		\STATE $I_2 = \frac{1}{N_2}\mathop{\sum}_{j\in\omega^{R}_i}w_{ij}q_j$, $N_2 = \mathop{\sum}_{j\in\omega^{R}_i}w_{ij}$
		\STATE $I_3 = \frac{1}{N_3}\mathop{\sum}_{j\in\omega^{U}_i}w_{ij}q_j$, $N_3 = \mathop{\sum}_{j\in\omega^{U}_i}w_{ij}$
		\STATE $I_4 = \frac{1}{N_4}\mathop{\sum}_{j\in\omega^{D}_i}w_{ij}q_j$, $N_4 = \mathop{\sum}_{j\in\omega^{D}_i}w_{ij}$
		\STATE $I_5 = \frac{1}{N_5}\mathop{\sum}_{j\in\omega^{NW}_i}w_{ij}q_j$, $N_5 = \mathop{\sum}_{j\in\omega^{NW}_i}w_{ij}$
		\STATE $I_6 = \frac{1}{N_6}\mathop{\sum}_{j\in\omega^{NE}_i}w_{ij}q_j$, $N_6 = \mathop{\sum}_{j\in\omega^{NE}_i}w_{ij}$
		\STATE $I_7 = \frac{1}{N_7}\mathop{\sum}_{j\in\omega^{SW}_i}w_{ij}q_j$, $N_7 = \mathop{\sum}_{j\in\omega^{SW}_i}w_{ij}$
		\STATE $I_8 = \frac{1}{N_8}\mathop{\sum}_{j\in\omega^{SE}_i}w_{ij}q_j$, $N_8 = \mathop{\sum}_{j\in\omega^{SE}_i}w_{ij}$
		\fi
		\STATE find $I_m$, such that $I_m = \mathop{\operatorname*{argmin}_{n \in S}{||q_i - I_n||_2^2}}$
		\ENSURE $I_m$
	\end{algorithmic}
\end{algorithm} 
%

By applying a filtering kernel $F$ in each side window, we can obtain eight outputs, denoted as $I_{i}^{'\theta,\rho}$, where $ \theta=k\times\frac{\pi}{2}, k\in[0,3]$ and $\rho\in\{0, r\}$
\begin{equation}
I_{i}^{'\theta,\rho,r} = F(q_i,\theta,\rho,r)
\end{equation}
To preserve the edges means that we want to minimize the distance between the input and the output at an edge, i.e., the filter output should be the same as or as close as possible to the input at an edge. Therefore, we choose the output of the side window that has the minimum $L_2$ distance to the input intensity as the final output, 
\begin{equation}
I'_{SWF} = \mathop{arg min}_{\forall I_{i}^{'\theta,\rho,r}}||q_{i}-I_{i}^{'\theta,\rho,r}||_2^2
\end{equation}
where $I'_{SWF}$ is the output of SWF. Eq. (4) is referred to as the SWF technique. Details of the procedure is described in Algorithm \ref{algo_swf}.
\begin{table}[htbp]
	\centering
	\caption{Summary of the output of BOX and S-BOX}
	\begin{tabular}{|c|c|c|}
		\hline
		Input&BOX&S-BOX\\
		\hline
		(a)&$\frac{(r+1)u+rv}{2r+1}$&$u$\\
		\hline
		(d)&$\frac{(r+1)u+rv}{2r+1}$&$u$\\
		\hline
		(g)&$\frac{(r+1)u+rv}{2r+1}$&$u$\\
		\hline
		(j)&$\frac{(r+1)^2u+((2r+1)^2-(r+1)^2)v}{(2r+1)^2}$&$u$\\
		\hline
		(m)&$u+\frac{r(r+1)\vartriangle{v}}{2(2r+1)}$&$u$\\
		\hline
		(p)&$v-\frac{r(r+1)\vartriangle{u}}{2r+1}$&$v-\frac{r}{2}\vartriangle{u}$\\
		\hline
	\end{tabular}
	\label{boxandsbox}
\end{table}
\begin{figure*}[htbp]
	\centering
	\subfigure[]{\includegraphics[height=0.89in]{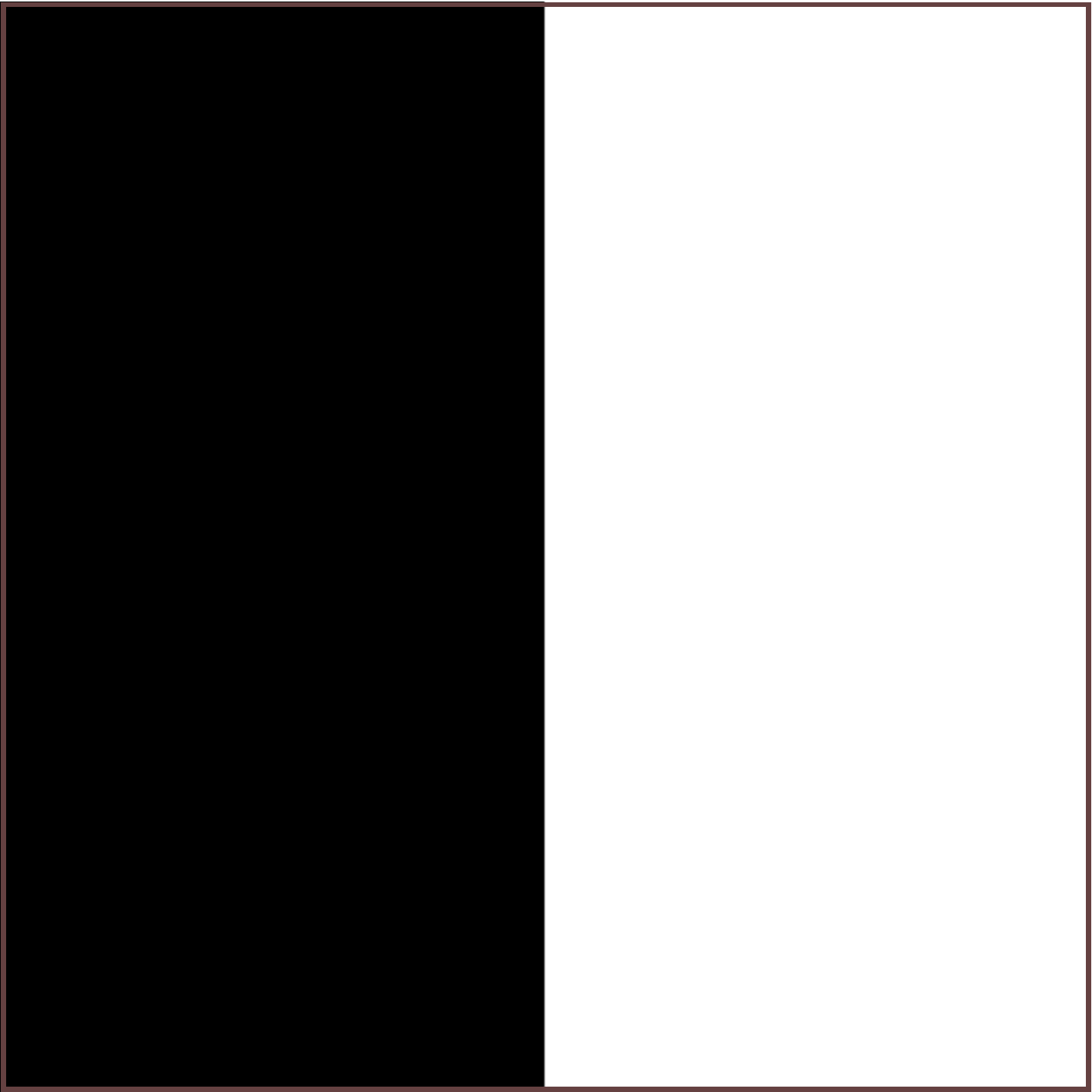}}
	\subfigure[]{\includegraphics[height=0.89in]{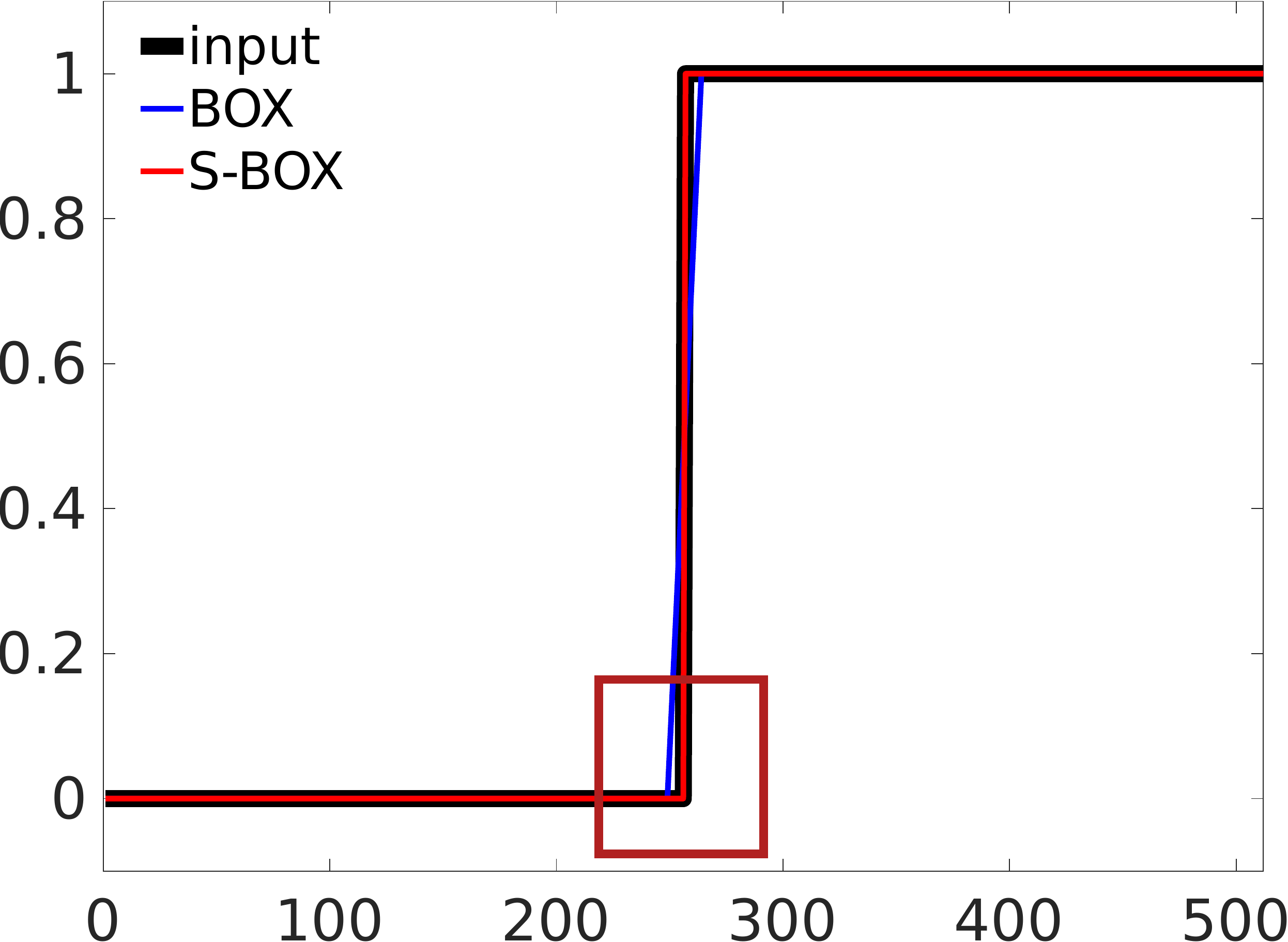}}
	\subfigure[]{\includegraphics[height=0.89in]{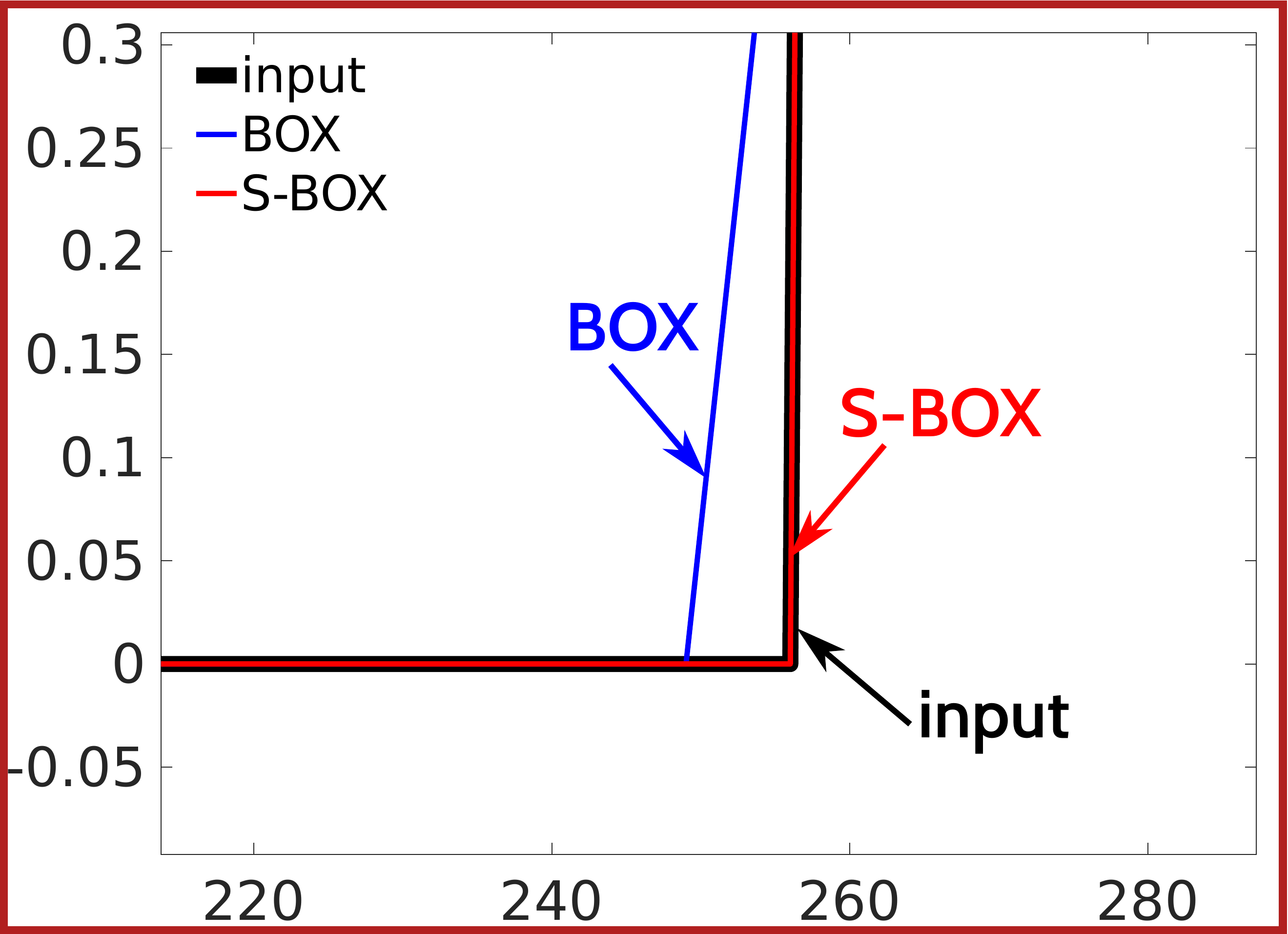}}
	\subfigure[]{\includegraphics[height=0.89in]{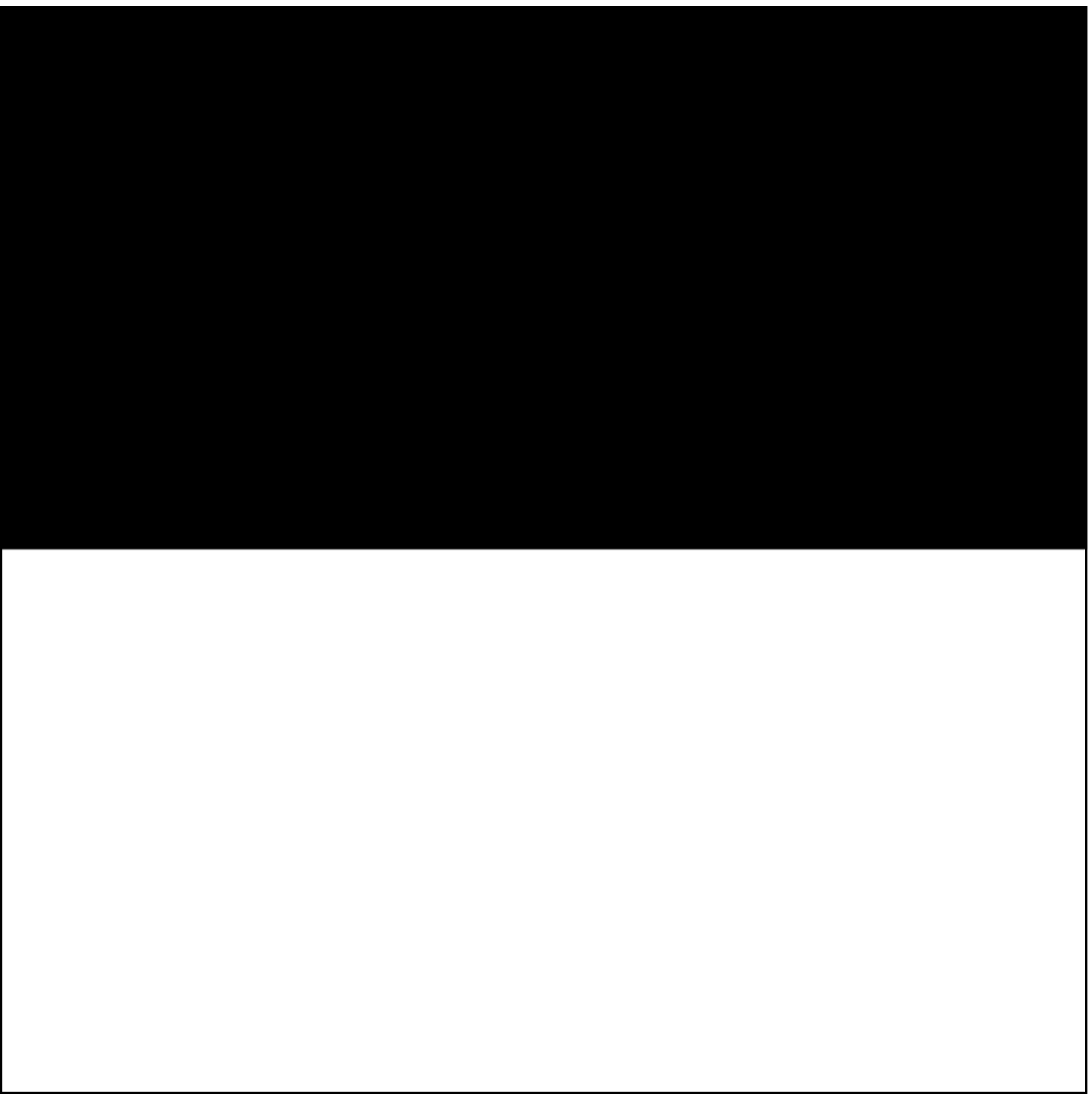}}
	\subfigure[]{\includegraphics[height=0.89in]{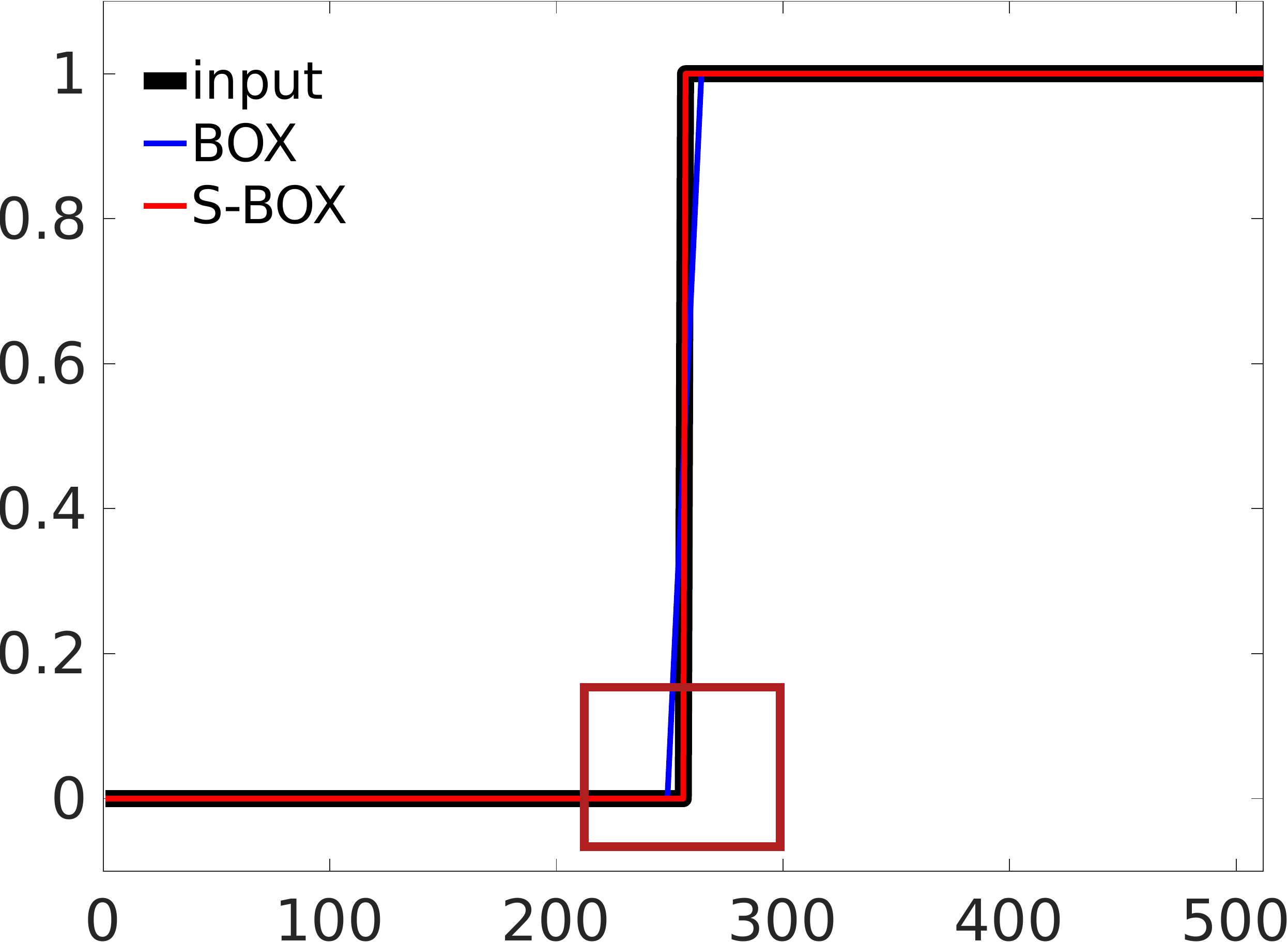}}
	\subfigure[]{\includegraphics[height=0.89in]{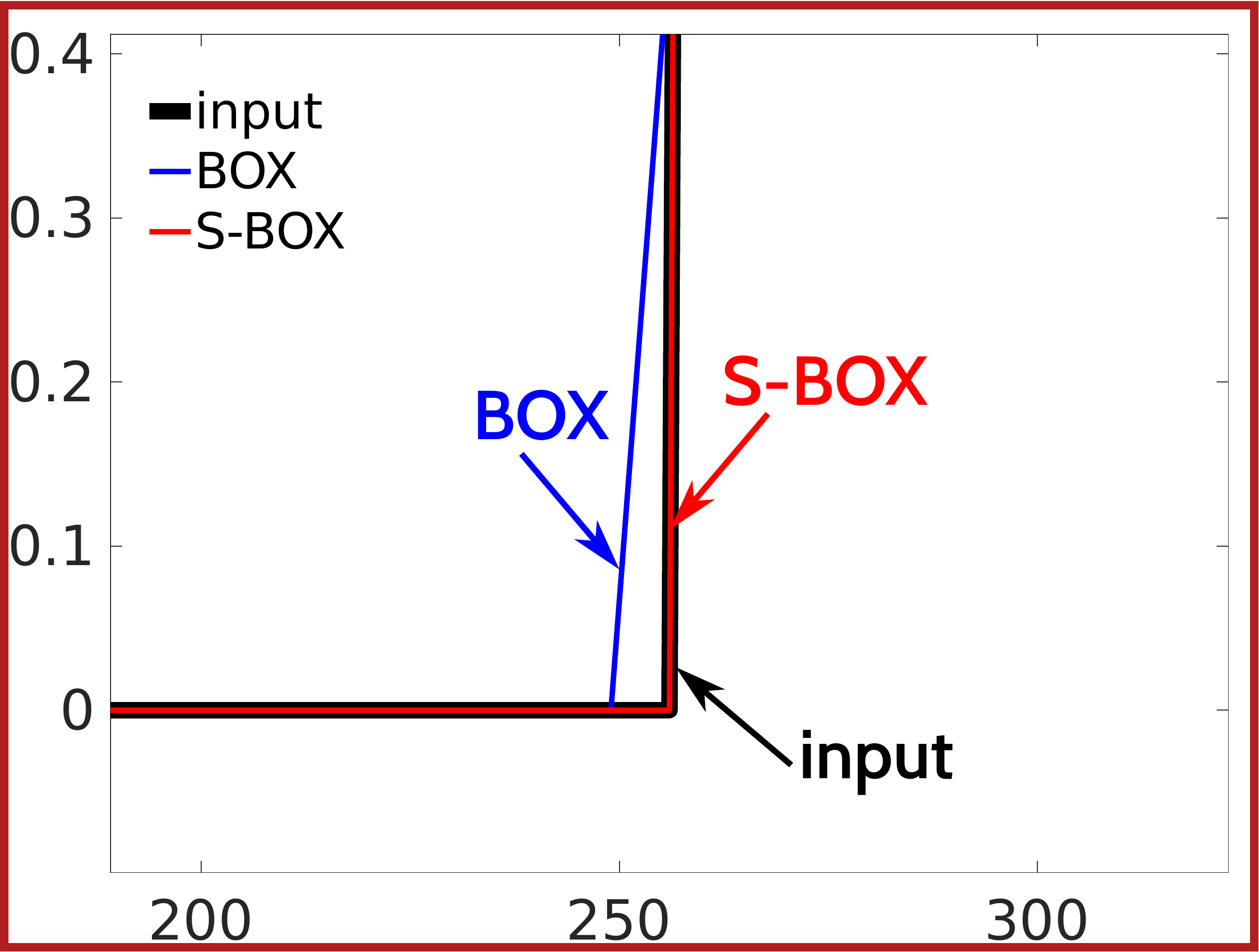}}
	\subfigure[]{\includegraphics[height=0.89in]{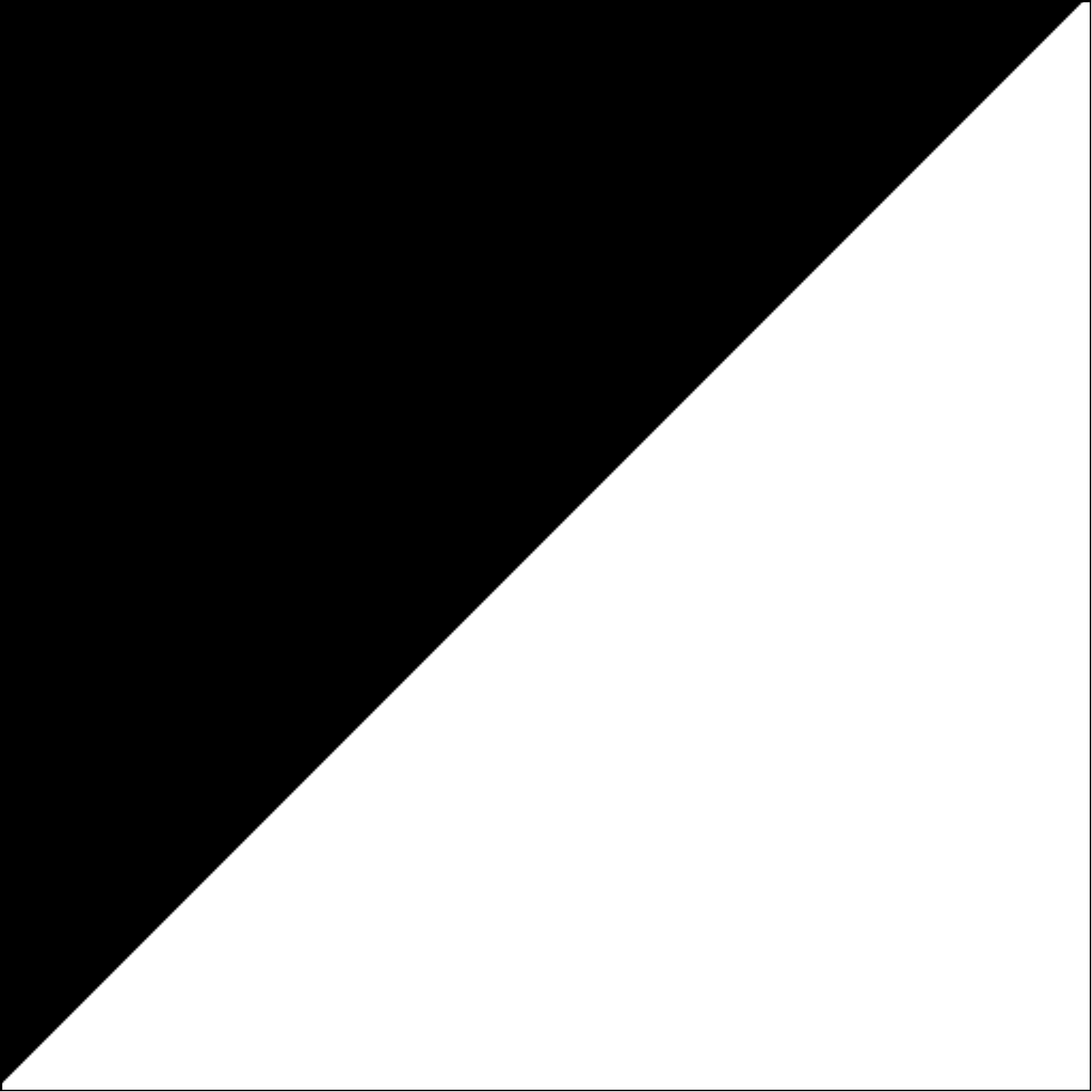}}
	\subfigure[]{\includegraphics[height=0.89in]{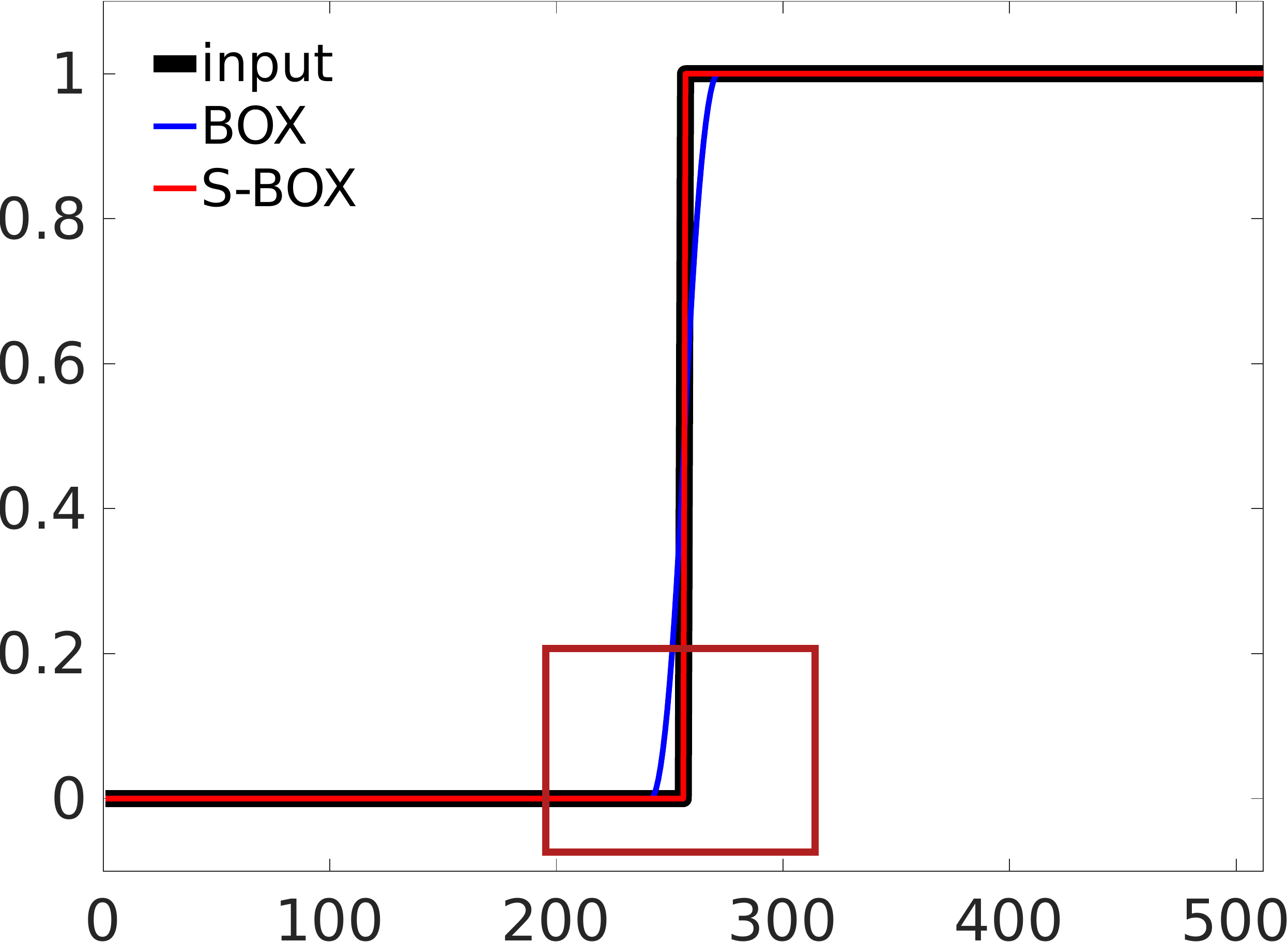}}
	\subfigure[]{\includegraphics[height=0.89in]{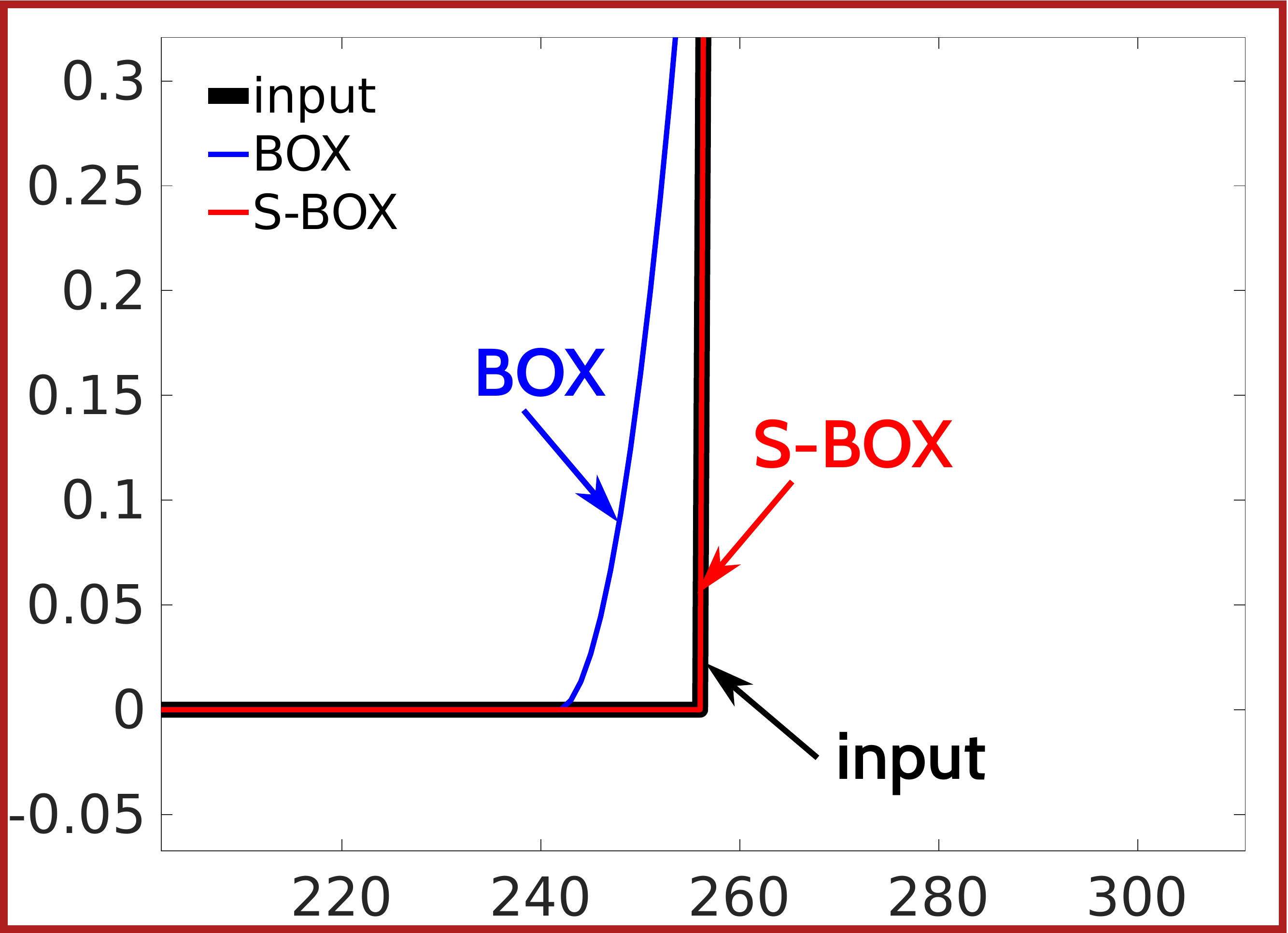}}
	\subfigure[]{\includegraphics[height=0.89in]{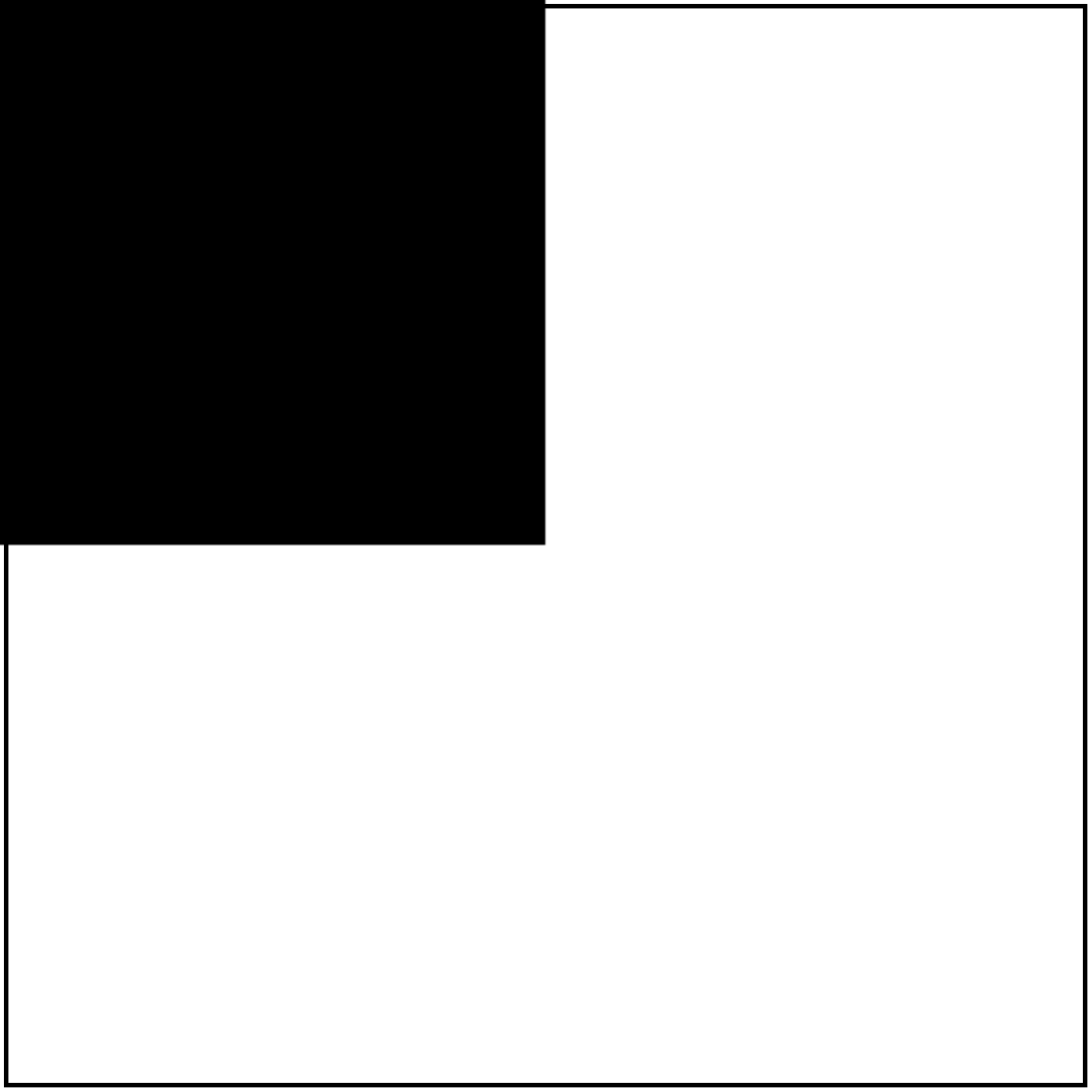}}
	\subfigure[]{\includegraphics[height=0.89in]{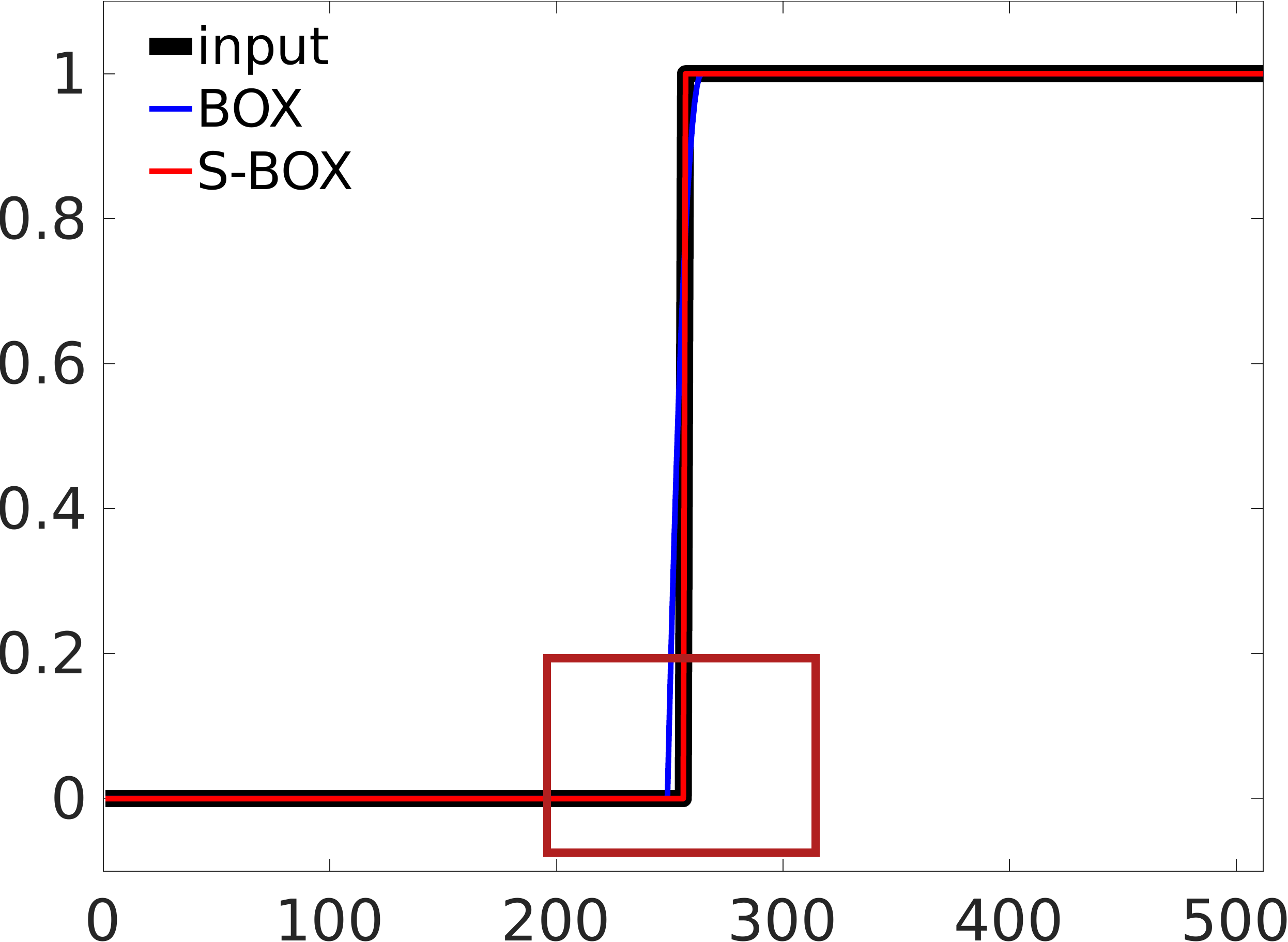}}
	\subfigure[]{\includegraphics[height=0.89in]{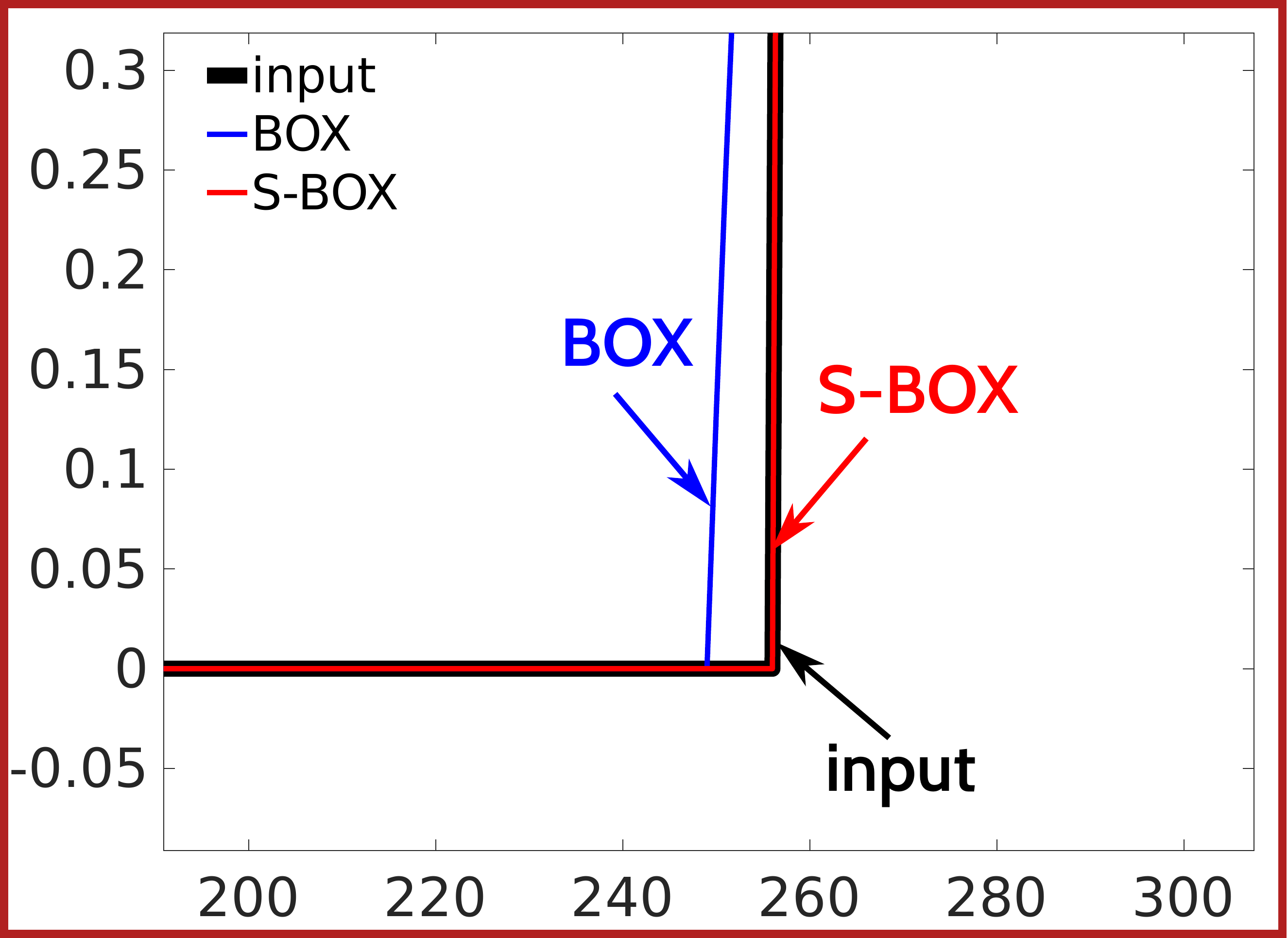}}
	\subfigure[]{\includegraphics[height=0.89in]{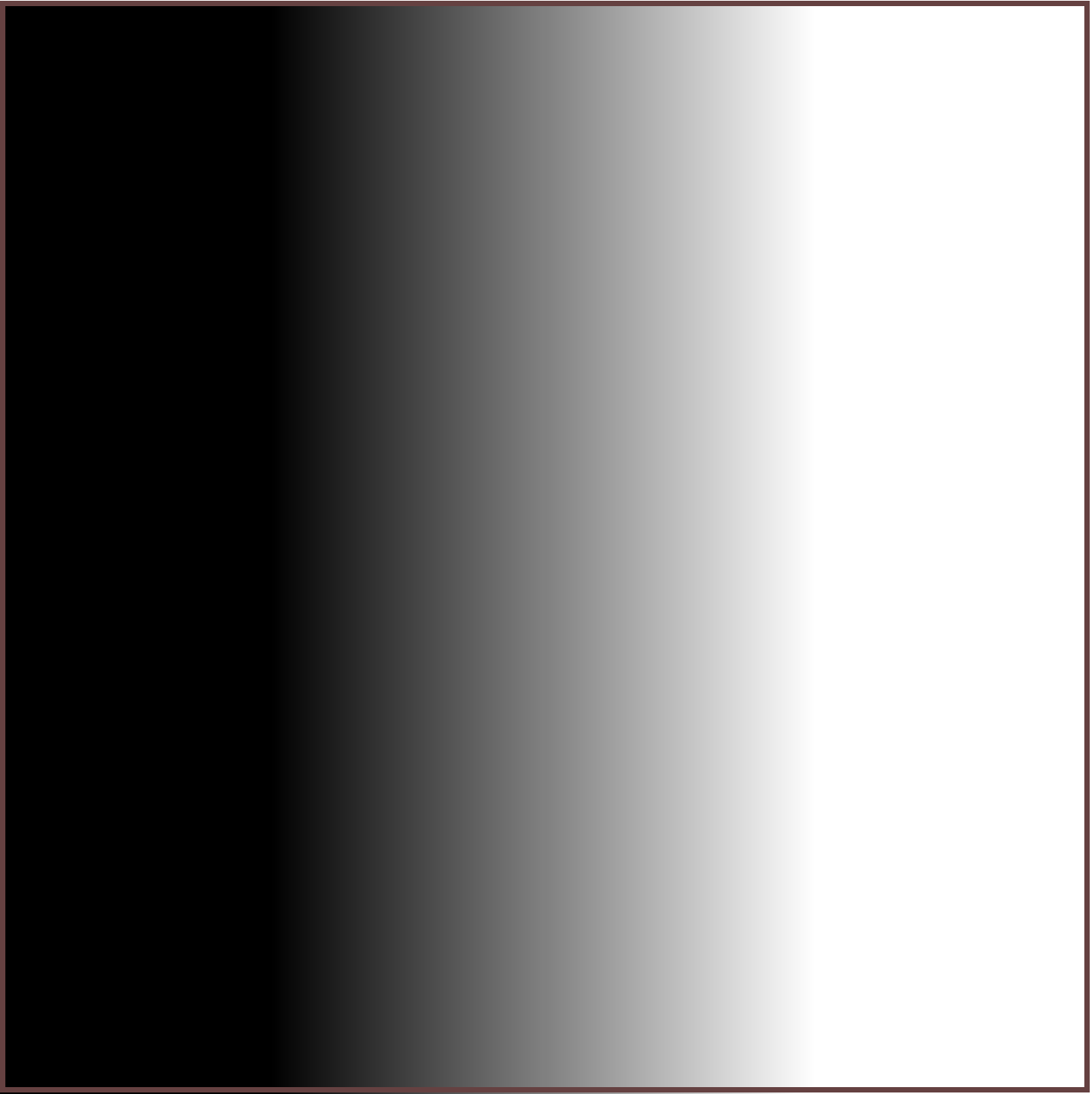}}
	\subfigure[]{\includegraphics[height=0.89in]{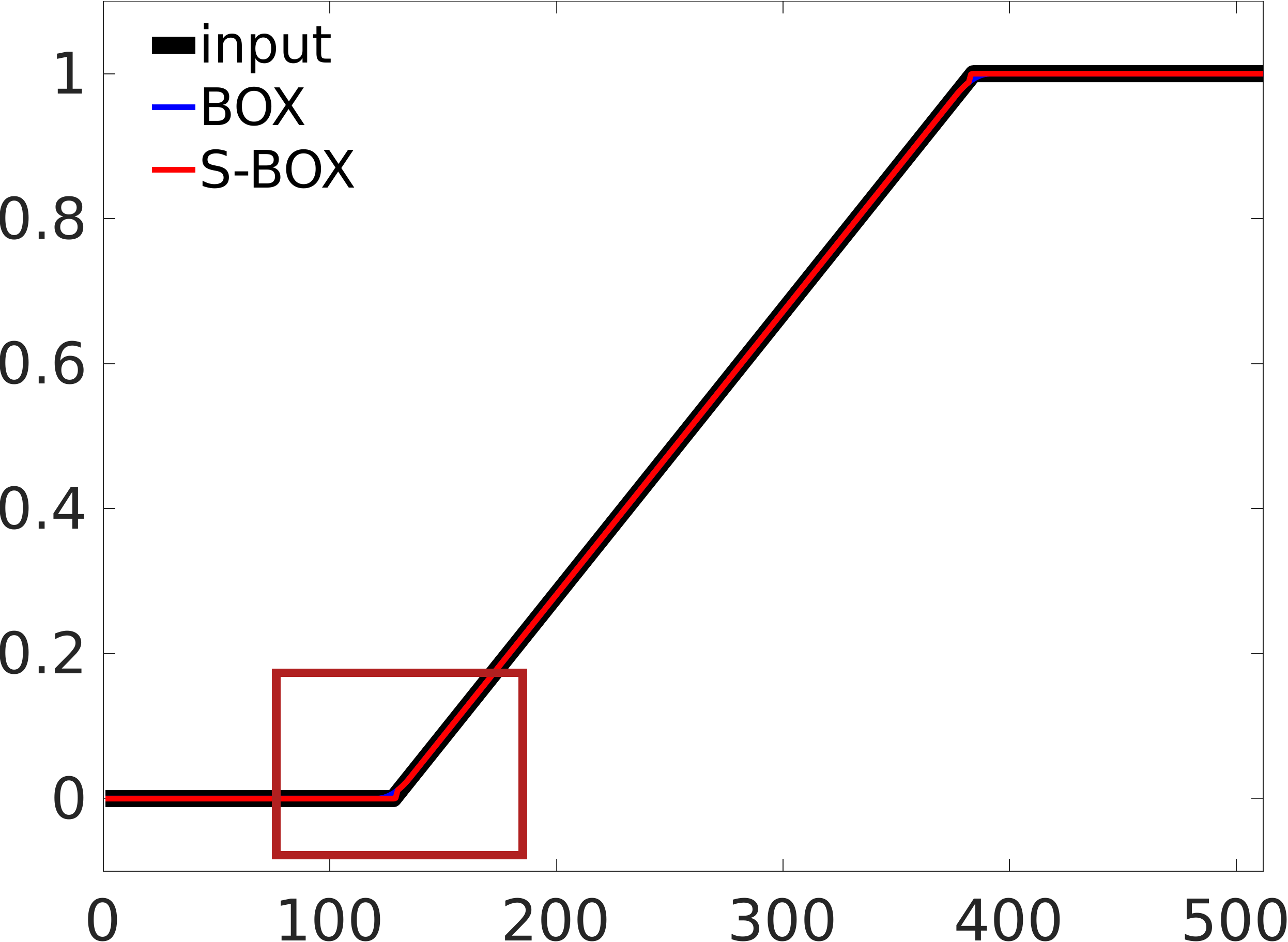}}
	\subfigure[]{\includegraphics[height=0.89in]{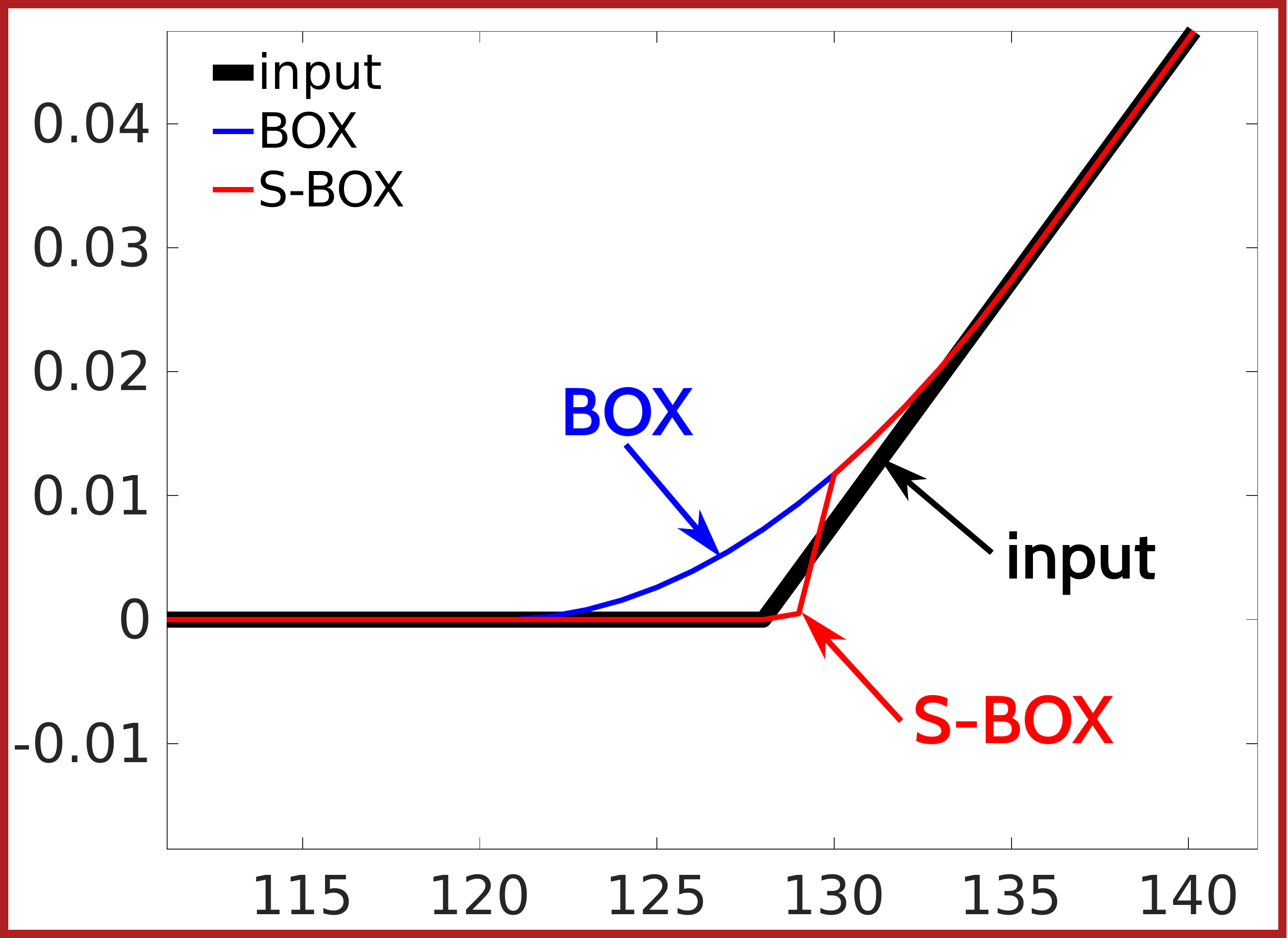}}
	\subfigure[]{\includegraphics[height=0.89in]{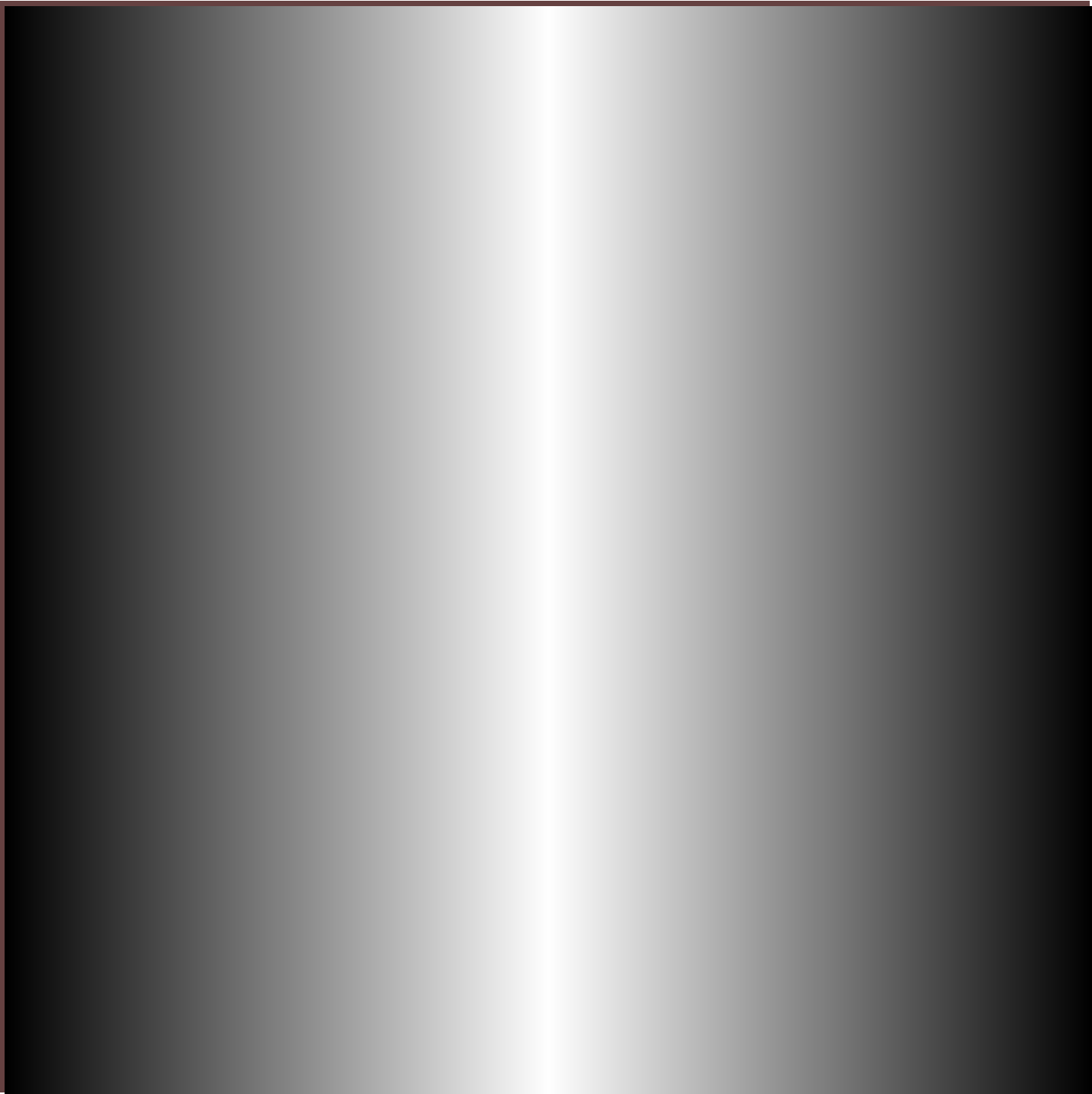}}
	\subfigure[]{\includegraphics[height=0.89in]{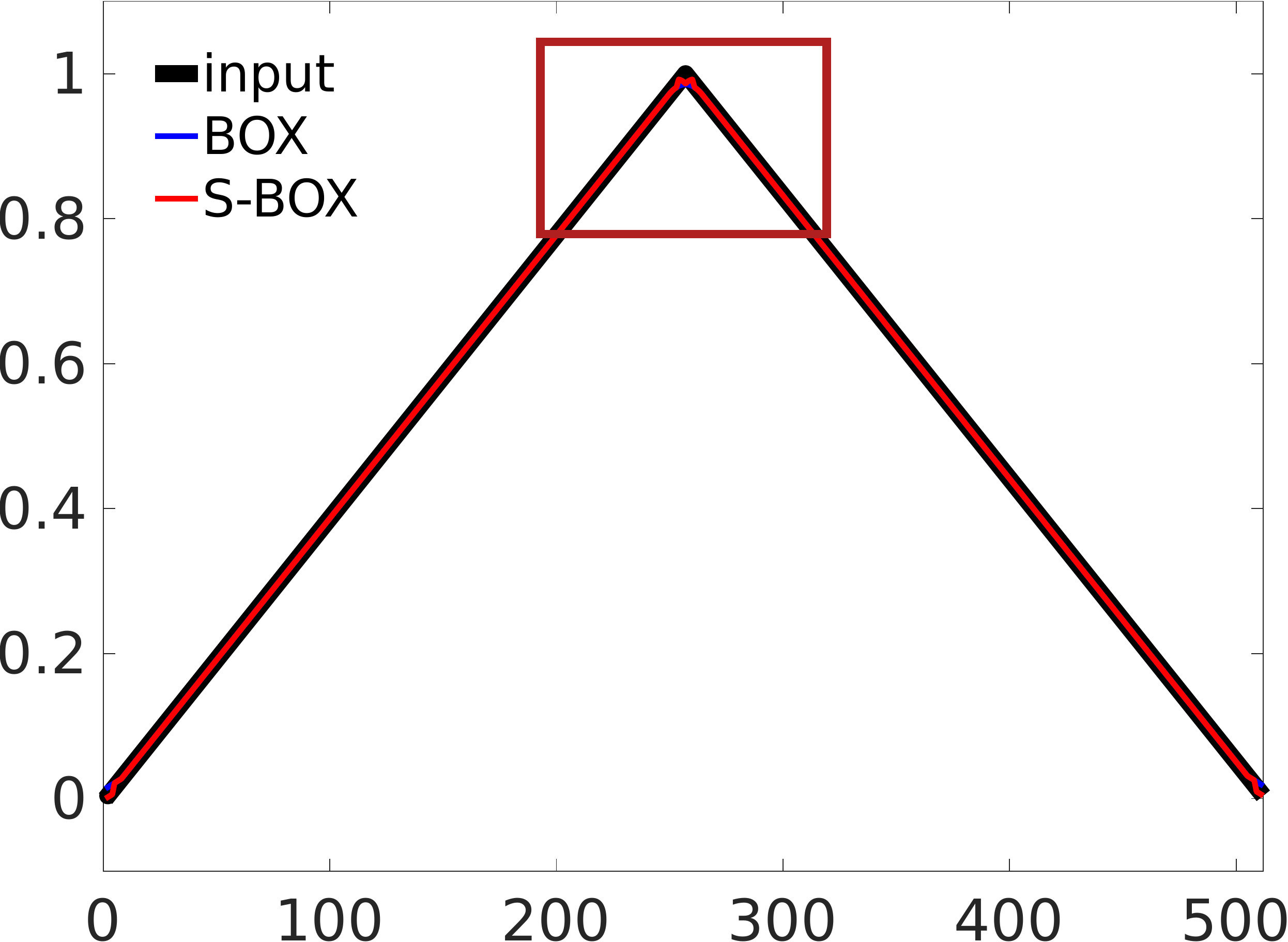}}
	\subfigure[]{\includegraphics[height=0.89in]{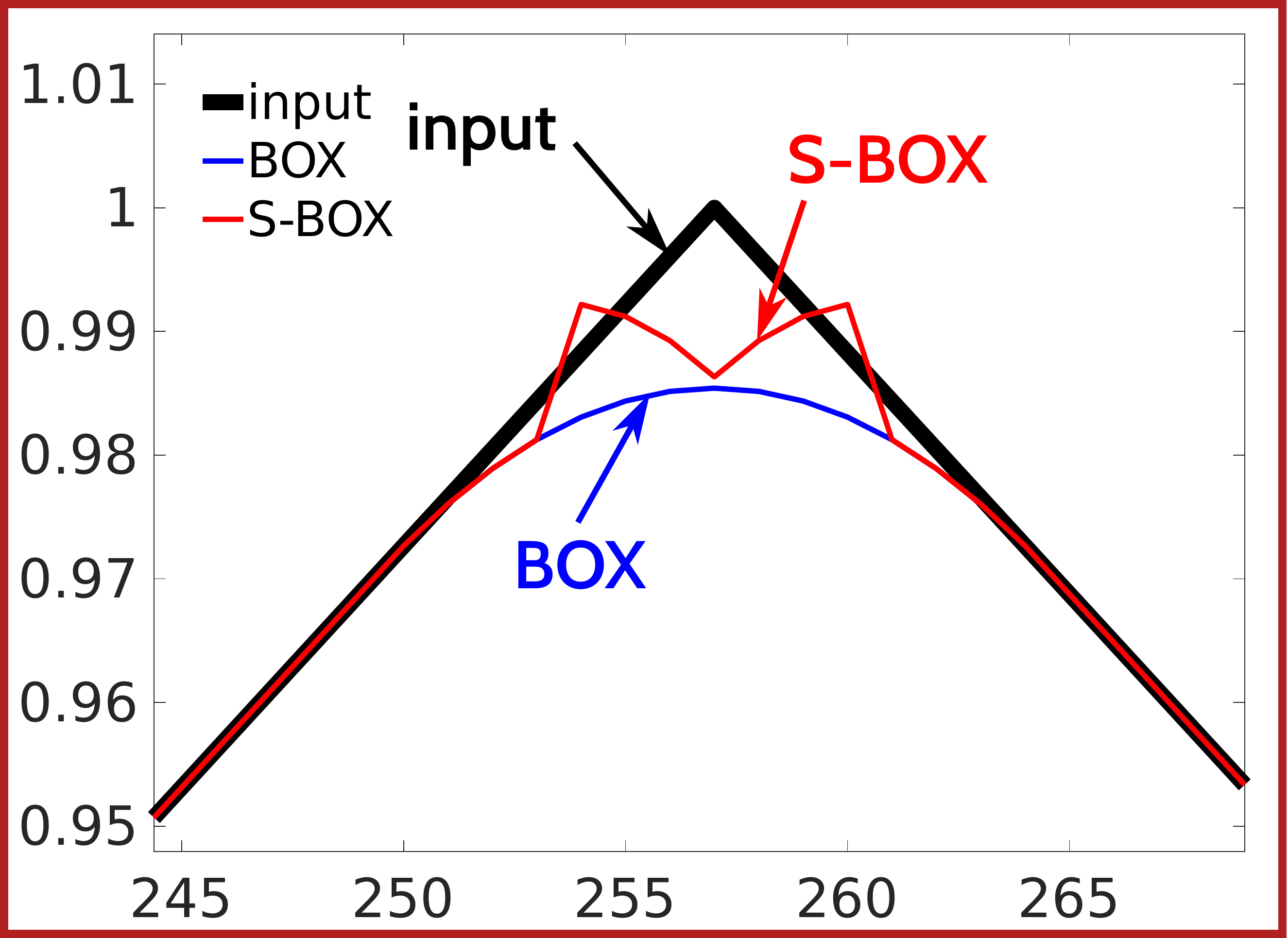}}
	\caption{
		\if false
		Comparing BOX and S-BOX on the testing images with different edges. (a) Testing image with a vertical edge. (b) One line of pixels extracted from the results in (a). (c) Zoomed in the red rectangle in (b). (d) Testing image with a horizontal edge. (e) One column of pixels extracted from the results in (d). (f) Zoomed in the red rectangle in (e). (g) Testing image with a diagonal edge. (h) One line of pixels extracted from the result in (g). (i) Zoomed in the red rectangle in (h). (j) Testing image with a corner. (k) The diagonal line of pixels extracted from the results in (j). (l) Zoomed in the red rectangle in (k). (m) Testing image with a ramp edge. (n) One line of pixels extracted from the results in (m). (o) Zoomed in the red rectangle in (n). (p) Testing image with a roof edge. (q) One line of pixels extracted from the results in (p). (r) Zoomed in the red rectangle in (q).
		\else
		Comparing BOX and S-BOX on the testing images with different edges. The first and forth columns (a), (g), (m), (d), (j) and (p) are input images with edge or corner. The second and fifth columns are middle line profiles for input, BOX filter and S-BOX filter. The third and sixth columns are the zoomed in region at the edge or corner location.
		\fi}
	\label{analysisfilters}
\end{figure*}
\begin{table*}
	{	\small
		\centering
		\caption{Summary of the output of each side window in S-BOX}
		\begin{tabular}{|c|c|c|c|c|c|c|c|c|c|c|}
			\hline
			Case&L&R&U&D&NW&NE&SW&SE\\
			\hline
			(a)&$\textbf{u}$&$\frac{u+rv}{r+1}$&$\frac{(r+1)u+rv}{2r+1}$&$\frac{(r+1)u+rv}{2r+1}$&$\textbf{u}$&$\frac{u+rv}{r+1}$&$\textbf{u}$&$\frac{u+rv}{r+1}$\\
			\hline
			(d)&$\frac{(r+1)u+rv}{2r+1}$&$\frac{(r+1)u+rv}{2r+1}$&$\textbf{u}$&$\frac{u+rv}{r+1}$&$\textbf{u}$&$\textbf{u}$&$\frac{u+rv}{r+1}$&$\frac{u+rv}{r+1}$\\
			\hline
			(g)&$\frac{(\frac{3r}{2}+1)u+\frac{r}{2}v}{2r+1}$&$\frac{(\frac{r}{2}+1)u+\frac{3r}{2}v}{2r+1}$&$\frac{(\frac{3r}{2}+1)u+\frac{r}{2}v}{2r+1}$&$\frac{(\frac{r}{2}+1)u+\frac{3r}{2}v}{2r+1}$&$\textbf{u}$&$\frac{(\frac{r}{2}+1)u+\frac{r}{2}v}{r+1}$&$\frac{(\frac{r}{2}+1)u+\frac{r}{2}v}{r+1}$&$\frac{((r+1)^2-1)v+u}{(r+1)^2}$\\
			\hline
			(j)&$\frac{(r+1)u+rv}{2r+1}$&$\frac{u+2rv}{2r+1}$&$\frac{(r+1)u+rv}{2r+1}$&$\frac{u+2rv}{2r+1}$&$\textbf{u}$&$\frac{u+rv}{r+1}$&$\frac{u+rv}{r+1}$&$\frac{((r+1)^2-1)v+u}{(r+1)^2}$\\
			\hline
			(m)&$\textbf{u}$&$u+\frac{r}{2}\vartriangle{v}$&$u+\frac{r(r+1)\vartriangle{v}}{2(2r+1)}$&$u+\frac{r(r+1)\vartriangle{v}}{2(2r+1)}$&$\textbf{u}$&$\frac{u}{r+1}+\frac{r}{2}\vartriangle{v}$&$\textbf{u}$&$\frac{u}{r+1}+\frac{r}{2}\vartriangle{v}$\\
			\hline
			(p)&$v-\frac{r}{2}\vartriangle{u}$&$v-\frac{r}{2}\vartriangle{u}$&$v-\frac{r(r+1)\vartriangle{u}}{2r+1}$&$v-\frac{r(r+1)\vartriangle{u}}{2r+1}$&{\tiny $v-\frac{r}{2}\vartriangle{u}$}&$v-\frac{r}{2}\vartriangle{u}$&$v-\frac{r}{2}\vartriangle{u}$&$v-\frac{r}{2}\vartriangle{u}$\\
			\hline
		\end{tabular}
		\label{outputsofsidewindow}
	}
\end{table*}

\subsection{Analysis of SWF}
In this section, we present a detailed analysis of the edge-preserving property of SWF technique. 
For analysis convenience, we use box filter (BOX) as an example and similar analysis can be performed on other forms of filter. This means that $F$ in eq. (3) is averaging and the resulting filter is called side window box filter (S-BOX).  

We compare the edge-preserving property of BOX and S-BOX filters. First of all, testing images with typical edges are generated, as shown in Fig.~\ref{analysisfilters}. There are six typical edges, including vertical edge (a), horizontal edge (d), diagonal edge (g), corner (j), ramp edge (m) and roof edge (p). For the vertical edge, horizontal edge, diagonal edge and corner, the pixel values for the black part of the edge is $u$ and the white part of the edge is $v$. For the ramp edge, the pixel values are increased from $u$ to $v$ with a step of $\vartriangle{v}$. For the roof edge, the top of the roof is $v$ and is decreased with a step of $\vartriangle{u}$. Based on these conditions, the outputs of BOX and S-BOX are deduced and the results are shown in Table~\ref{boxandsbox}. From Table~\ref{boxandsbox}, we can easily see that S-BOX better preserves the edges in (a)$\sim$(m) than BOX. It is also easy to prove $\frac{r}{2}<\frac{r(r+1)}{2r+1}$, so S-BOX can better preserve the roof edge than BOX, too. 


In order to observe the details of the edge-preserving property of each side window in S-BOX, the output of each side window is shown in Table~\ref{outputsofsidewindow}. The results which preserve the edges are shown in bold. We can make the following observations: 
\if false
\begin{figure}[h]
	\centering
	\subfigure[]{
		\begin{minipage}{1.03in}
			\centering
			\includegraphics[height=0.5in]{figures/wholegausskernel.png}
			\includegraphics[height=0.5in]{figures/fullkernel1.png}
	\end{minipage}}
	\subfigure[]{
		\begin{minipage}{1.03in}
			\centering
			\includegraphics[height=0.5in]{figures/halfgausskernel.png}
			\includegraphics[height=0.5in]
			{figures/halfkernel1.png}
	\end{minipage}}
	\subfigure[]{
		\begin{minipage}{1.03in}
			\centering
			\includegraphics[height=0.5in]{figures/wngausskernel.png}
			\includegraphics[height=0.5in]{figures/southeastkernel1.png}
	\end{minipage}}
	\caption{Definition of kernel in GAU ans S-GAU. (a) The kernel in GAU and its mapping on $z=0$ plane (b) The kernel in $R$ side window of S-GAU and its mapping on $z=0$ plane (c) The kernel in $SE$ side window of S-GAU and its mapping on $z=0$ plane. The red and blue curves are the mapping of the kernels on $x=0$ and $y=0$ plane.}
	\label{sgau}
\end{figure}
\fi
\begin{figure*}[ht]
	\centering
	\subfigure[Input]{
		\begin{minipage}{1.08in}
			\centering
			\includegraphics[width=1.08in]{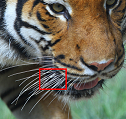}
			\includegraphics[width=0.52in]{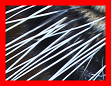}
	\end{minipage}}
	\subfigure[BOX vs. S-BOX]{
		\begin{minipage}{1.08in}
			\centering
			\includegraphics[width=1.08in]{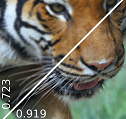}
			\includegraphics[width=0.52in]{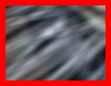}
			\includegraphics[width=0.52in]{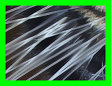}
	\end{minipage}}
	\subfigure[GAU vs. S-GAU]{
		\begin{minipage}{1.08in}
			\centering
			\includegraphics[width=1.08in]{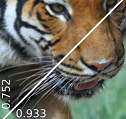}
			\includegraphics[width=0.52in]{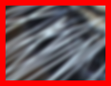}
			\includegraphics[width=0.52in]{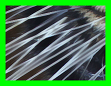}
	\end{minipage}}
	\subfigure[MED vs. S-MED]{
		\begin{minipage}{1.08in}
			\centering
			\includegraphics[width=1.08in]{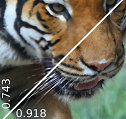}
			\includegraphics[width=0.52in]{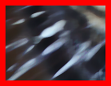}
			\includegraphics[width=0.52in]{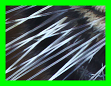}
	\end{minipage}}
	\subfigure[BIL vs. S-BIL]{
		\begin{minipage}{1.08in}
			\centering
			\includegraphics[width=1.08in]{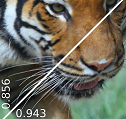}
			\includegraphics[width=0.52in]{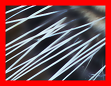}
			\includegraphics[width=0.52in]{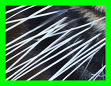}
	\end{minipage}}
	\subfigure[GUI vs. S-GUI]{
		\begin{minipage}{1.08in}
			\centering
			\includegraphics[width=1.08in]{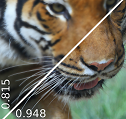}
			\includegraphics[width=0.52in]{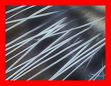}
			\includegraphics[width=0.52in]{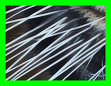}
	\end{minipage}}
	\caption{Image smoothing ($r=7$, $\sigma=4$ for GAU and S-GAU, $\sigma_s = 7, \sigma_r = 0.3$ for BIL and S-BIL, $\epsilon=0.1$ for GUI and S-GUI). The upper left part of each result is from the traditional filter and the zoomed in patch is with red rectangle. The lower right part of each result is from the side window version and the zoomed in patch is with green rectangle. The number shown on each image is the SSIM\cite{ssim} value.}
	\label{grayimagesmoothing}
\end{figure*}

\begin{itemize}
	\item the $L, NW, SW$ side windows can preserve the edges on the left of the vertical edge. It is easy to deduce that the $R, NE, SE$ side windows can preserve the edges on the right of the vertical edge.
	\item the $U, NW, NE$ side windows can preserve the edges above the horizontal edge. Again, it is easy to deduce that the $D, SW, SE$ side windows can preserve the edges below the horizontal edge.
	\item the $NW$ side window can preserve the edges above the diagonal edge and on the corner. It is easy to deduce that the $NE, SW, SE$ side windows can preserve the diagonal edges and corner with other directions.
	\item the $L, NW, SW$ side windows can preserve the ramp edge.
	\item although the side windows can not preserve the roof edge completely, seven of them have better results than BOX.
\end{itemize}

We also show the experimental results in Fig.~\ref{analysisfilters}. In the experiments, we set $u=0,v=1$ and $r=7$. One line (or column or diagonal) of pixels is extracted from each result and zoomed in. The results are consistent with the theoretical deduction. Visually, the sharp edges are smoothed away by BOX while preserved very well by S-BOX.
\begin{figure*}[ht]
	\centering
	\subfigure[Input]{
		\begin{minipage}{1.08in}
			\centering
			\includegraphics[width=1.08in]{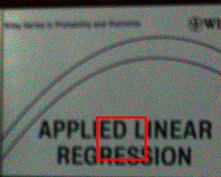}
			\includegraphics[width=0.52in]{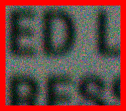}
	\end{minipage}}
	\subfigure[BOX vs. S-BOX]{
		\begin{minipage}{1.08in}
			\centering
			\includegraphics[width=1.08in]{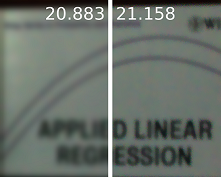}
			\includegraphics[width=0.52in]{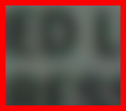}
			\includegraphics[width=0.52in]{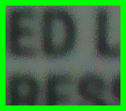}
	\end{minipage}}
	\subfigure[GAU vs. S-GAU]{
		\begin{minipage}{1.08in}
			\centering
			\includegraphics[width=1.08in]{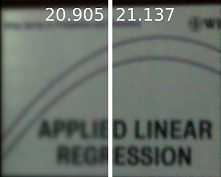}
			\includegraphics[width=0.52in]{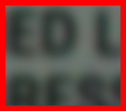}
			\includegraphics[width=0.52in]{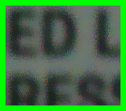}
	\end{minipage}}
	\subfigure[MED vs. S-MED]{
		\begin{minipage}{1.08in}
			\centering
			\includegraphics[width=1.08in]{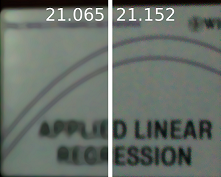}
			\includegraphics[width=0.52in]{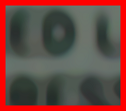}
			\includegraphics[width=0.52in]{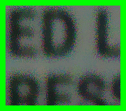}
	\end{minipage}}
	\subfigure[BIL vs. S-BIL]{
		\begin{minipage}{1.08in}
			\centering
			\includegraphics[width=1.08in]{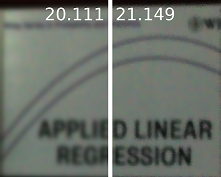}
			\includegraphics[width=0.52in]{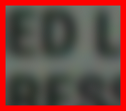}
			\includegraphics[width=0.52in]{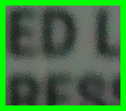}
	\end{minipage}}
	\subfigure[GUI vs. S-GUI]{
		\begin{minipage}{1.08in}
			\centering
			\includegraphics[width=1.08in]{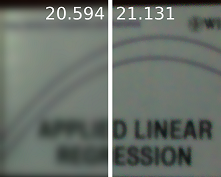}
			\includegraphics[width=0.52in]{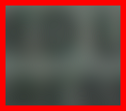}
			\includegraphics[width=0.52in]{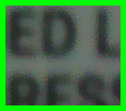}
	\end{minipage}}
	\caption{Image denoising ($r=10, \sigma=5$ for GAU and S-GAU, $\sigma_s = 10, \sigma_r = 0.3$ for BIL and S-BIL, $\epsilon=0.1$ for GUI and S-GUI, $iteration=5$ ). The left part of each result is from the traditional filter and the zoomed in patch is with red rectangle. The right part of each result is from the side window version and the zoomed in patch is with green rectangle. The number shown on each image is PSNR.}
	\label{imagedenoising}
\end{figure*}
\section{Popular Filters under the SWF Framework}
By changing $F$ to other kernels, one can easily embed the side window technique into other filters. In this section, we will discuss how to embed side window technique into Gaussian filter, median filter, bilateral filter and guided filter. To simplify expression, the filters' name are abbreviated by their first three letters and their SWF versions are abbreviated by adding another `S-'. For example, the Gaussian filter and side window Gaussian filter are abbreviated as GAU and S-GAU, respectively. 

In S-GAU, $F$ is a half of or a quarter of the  Gaussian kernel. 
Because the kernel of GAU crosses over the potential edges, it may blur the edges. By contrast, the kernel of S-GAU alleviates this problem so it can better preserve the edges.

In S-MED, $F$ is the operation of calculating the median value. Since the output has the minimal distance from the input intensity, S-MED can better preserve the edges than MED. That is, it selects a window under which the median of the pixels is closest to the input. 

In S-BIL, the kernel is calculated based on the geometric closeness and photometric similarity as in BIL. Since S-BIL can prevent the diffusion from crossing the edges, it can improve the edge-preserving property of BIL. 

GUI averages the values of the parameters in all the windows that cover the target pixel. Again, this operation may blur potential edges. To avoid this problem, S-GUI ensures that the side windows do not cross over the target pixel. It slides each side window along its side on which the target pixel is located until the target pixel is outside of the side window. In this way, $2r+1$ sliding windows are obtained for the $L, R, U, D$ side window and averaged to obtain the outputs of these side windows. For the $NW,NE,SW, SE$ side windows, sliding can only get one output for each. The final output is chosen according to eq. (4).
\begin{figure}[ht]
	\centering
	\subfigure[Input]{
		\begin{minipage}{1.02in}
			\centering
			\includegraphics[width=1in]{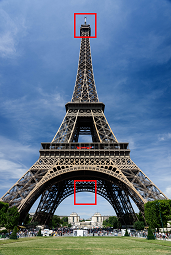}
			\includegraphics[width=0.49in]{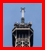}
			\includegraphics[width=0.49in]{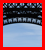}
	\end{minipage}}
	\subfigure[BIL]{
		\begin{minipage}{1.02in}
			\centering
			\includegraphics[width=1in]{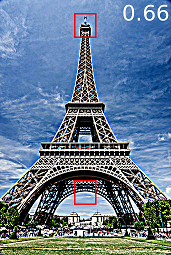}
			\includegraphics[width=0.49in]{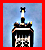}
			\includegraphics[width=0.49in]{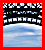}
	\end{minipage}}
	\subfigure[GUI]{
		\begin{minipage}{1.02in}
			\centering
			\includegraphics[width=1in]{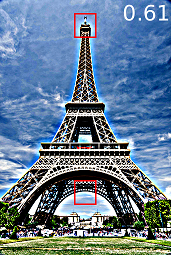}
			\includegraphics[width=0.49in]{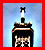}
			\includegraphics[width=0.49in]{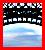}
	\end{minipage}}\\
	\hspace{75px}
	\subfigure[S-BIL]{
		\begin{minipage}{1.02in}
			\centering
			\includegraphics[width=0.49in]{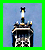}
			\includegraphics[width=0.49in]{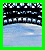}
			\includegraphics[width=1in]{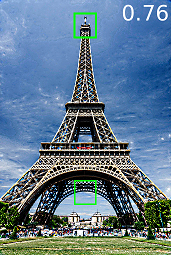}
	\end{minipage}}
	\subfigure[S-GUI]{
		\begin{minipage}{1.02in}
			\centering
			\includegraphics[width=0.49in]{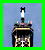}
			\includegraphics[width=0.49in]{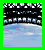}
			\includegraphics[width=1in]{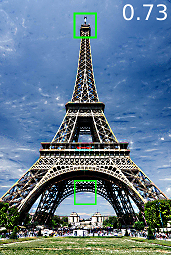}
	\end{minipage}}
	\caption{Image enhancement ($\sigma_s = 7, \sigma_r = 0.3$ for BIL and S-BIL, $r = 7, \epsilon = 0.1$ for GUI and S-GUI). The number shown on each image is the SSIM value.}
	\label{imageenhancement}
\end{figure}

\begin{figure}[ht]
	\centering
	\subfigure[BIL]{
		\begin{minipage}{1.52in}
			\centering
			\includegraphics[width=1.52in]{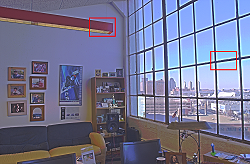}
			\includegraphics[width=0.73in]{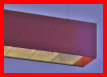}
			\includegraphics[width=0.73in]{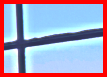}
	\end{minipage}}
	\subfigure[GUI]{
		\begin{minipage}{1.52in}
			\centering
			\includegraphics[width=1.52in]{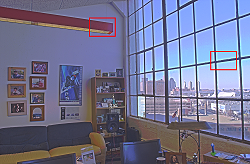}
			\includegraphics[width=0.73in]{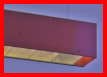}
			\includegraphics[width=0.73in]{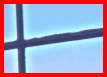}
	\end{minipage}}\\
	\subfigure[S-BIL]{
		\begin{minipage}{1.52in}
			\centering
			\includegraphics[width=0.73in]{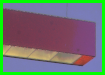}
			\includegraphics[width=0.73in]{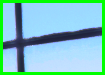}
			\includegraphics[width=1.52in]{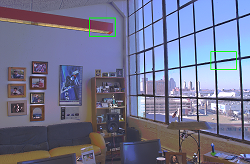}
	\end{minipage}}
	\subfigure[S-GUI]{
		\begin{minipage}{1.52in}
			\centering
			\includegraphics[width=0.73in]{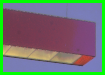}
			\includegraphics[width=0.73in]{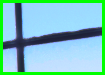}
			\includegraphics[width=1.52in]{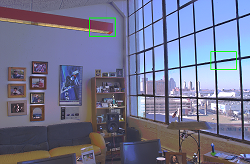}
	\end{minipage}}
	\caption{HDR tone mapping ($\sigma_s = 5, \sigma_r = 0.3$ for BIL and S-BIL, $r = 5, \epsilon = 0.1$ for GUI and S-GUI).}
	\label{hdr}
\end{figure}

\begin{figure*}[ht]
	\centering
	\subfigure[Input]{
		\begin{minipage}{1.2in}
			\centering
			\includegraphics[width=1.2in]{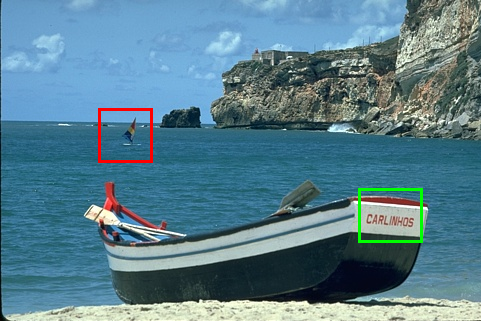}
			\includegraphics[width=0.52in]{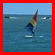}
			\includegraphics[width=0.6in]{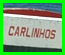}
	\end{minipage}}
	\subfigure[RTV]{
		\begin{minipage}{1.2in}
			\centering
			\includegraphics[width=1.2in]{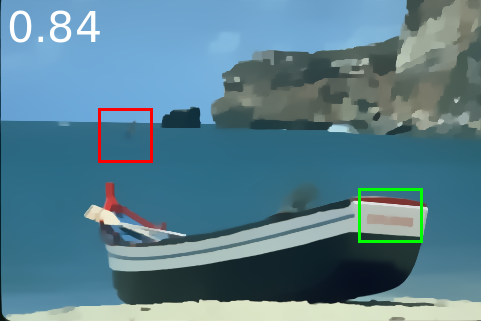}
			\includegraphics[width=0.52in]{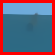}
			\includegraphics[width=0.6in]{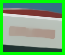}
	\end{minipage}}
	\subfigure[RGF]{
		\begin{minipage}{1.2in}
			\centering
			\includegraphics[width=1.2in]{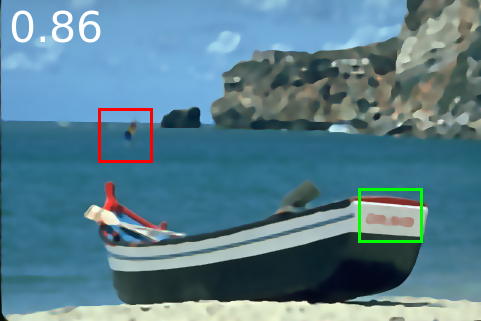}
			\includegraphics[width=0.52in]{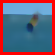}
			\includegraphics[width=0.6in]{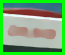}
	\end{minipage}}
	\subfigure[IWGF]{
		\begin{minipage}{1.2in}
			\centering
			\includegraphics[width=1.2in]{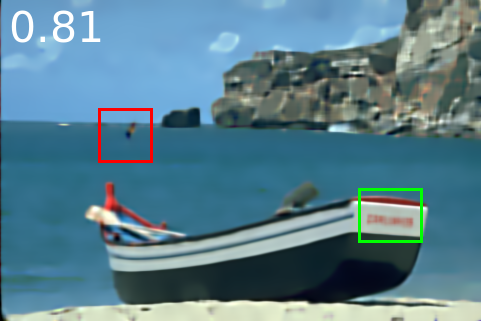}
			\includegraphics[width=0.52in]{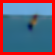}
			\includegraphics[width=0.6in]{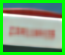}
	\end{minipage}}
	\subfigure[IS-WGF]{
		\begin{minipage}{1.2in}
			\centering
			\includegraphics[width=1.2in]{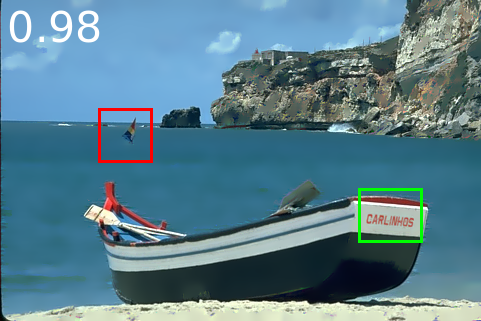}
			\includegraphics[width=0.52in]{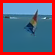}
			\includegraphics[width=0.6in]{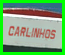}
	\end{minipage}}
	\caption{Structure-preserving and texture-removing on natural image from BSR Dataset ($r=5$, $\epsilon=0.005$, $iteration = 10$, $\lambda = 100$). The number shown on each image is the SSIM value.}
	\label{bsrdataset}
\end{figure*}

\section{Applications}
In this section, the side window technique is applied to various image processing applications and its performance is compared with traditional filters and methods. More results are presented in the Supplement. The images are best viewed electronically on a high resolution monitor.
\subsection{Image smoothing}
Fig.~\ref{grayimagesmoothing} shows the smoothing results of the filters on an image \cite{ntire}. The upper left part of each result is from the traditional filter and the zoomed in patch is with red rectangle. The lower right part of each result is from the side window version and the zoomed in patch is with green rectangle. As can be seen, the corresponding side window filter outperforms the original filter in preserving edges. This is more clearly shown in the zoomed in patches that the side window filters can better preserve the tiger's whiskers. When comparing the non-linear filters, this improvement is also obvious. This means that the side window technique can also improve the edge-preserving property of non-linear filters. This shows the potential for the side window technique to be widely used in more applications.
\subsection{Image denoising}
Fig.~\ref{imagedenoising} shows the results of iteratively applying different filters to remove noise of a low light image \cite{renoir}. The left part of each result is from the traditional filter and the zoomed in patch is with red rectangle. The right part of each result is from the side window version and the zoomed in patch is with green rectangle. BOX, GAU, MED, BIL and GUI remove the noises but blur the edges at the same time. On the other hand, the side window version of these filters can preserve edges and remove noises at the same time. These results further demonstrate the excellent edge preserving property of the new side window technique. 
\subsection{Image enhancement}
Image enhancement is often performed in image processing \cite{gf}\cite{wgf}. An enhanced image can be obtained by 
\begin{equation}
Enhanced = q + \alpha \times (q-I^{'})
\end{equation}
where $\alpha$ is an amplification parameter and is fixed to 5 in all the experiments in this section. An example of image enhancement result is shown in Fig.~\ref{imageenhancement}. From the zoomed in patches, we can see that the halo artifacts exist along the edges in the results of the filters without implementing the side window technique. However, the artifacts have disappeared in the results of the side window versions of the filters. This can once again be attributed to the excellent edge-preserving property of the side window technique. 
\subsection{HDR tone mapping}
In \cite{fbf} a technique based on bilateral filter was proposed for displaying HDR images, which reduces the dynamic range while preserving detail. Briefly the operation works as follows:
\begin{equation}
I' = \gamma \times q_{b} + q_{d}
\end{equation}
$q_{b}$ is the bilateral filter output, $q_d$ is the difference between the original HDR radiance map and $q_{b}$, and $\gamma\in(0,1)$ is the compression factor which determines the scale of dynamic range compression. For specific details please refer to \cite{fbf} and their companion website \cite{eh3}. In this experiment, we replace BIL and GUI by their side window versions. 

Fig.~\ref{hdr} shows examples of the results. From the zoomed in patches, we can see that halo artifacts exist in the results of filters without side window technique, while they do not exist in the results of their side window filters. These results can once again be attributed to the good edge-preserving property of SWF.
\subsection{Structure-preserving and texture-removing on natural image}
The goal of this application is to extract image structures and remove the textures \cite{rtv}. The side window technique is embedded into the weighted guided filter (WGF) \cite{wgf} to form a new filter called S-WGF. By combining WGF and S-WGF in the iteration framework of \cite{rgf}, we propose a new structure-preserving texture-removing filter, termed iterated side window weighted guided filter (IS-WGF). In this filter, an edge-aware weight \cite{wgf} and a threshold $\lambda$ is used to easily distinguish structures from textures. The structures are preserved by S-WGF and textures are removed by WGF. 

IS-WGF is compared with iterated weighted guided filter (IWGF), relative total variance (RTV) \cite{rtv} and rolling guidance filter (RGF) \cite{rgf}. IWGF is obtained by combining WGF with the iterative framework of \cite{rgf}, RTV is the state of art of this application and RGF is the original algorithm in \cite{rgf}. They are applied to smooth natural images with obvious textures and structures, as shown in Fig.~\ref{bsrdataset}(a). It is chosen from the BSR Dataset \cite{bsr}. The wave in the sea is viewed as textures and the sailing boat on the sea is viewed as structures. From the results we can see that only IS-WGF can preserve the structures while all other filters fail. This example demonstrates the excellent structure-preserving property of side window technique.

\subsection{Mutual structure extraction}
\begin{figure}[!htbp]
	\centering
	\subfigure[Input]{
		\begin{minipage}{1.02in}
			\centering
			\includegraphics[width=1.02in]{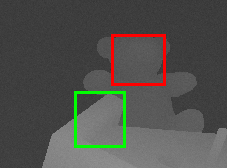}
			\includegraphics[width=0.53in]{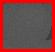}
			\includegraphics[width=0.455in]{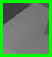}
			\includegraphics[width=1.02in]{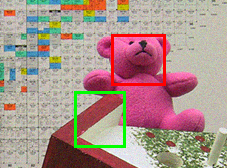}
			\includegraphics[width=0.53in]{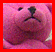}
			\includegraphics[width=0.455in]{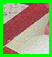}
	\end{minipage}}
	\subfigure[MJF]{
		\begin{minipage}{1.02in}
			\centering
			\includegraphics[width=1.02in]{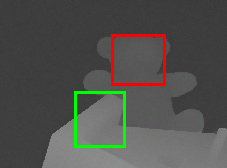}
			\includegraphics[width=0.53in]{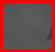}
			\includegraphics[width=0.455in]{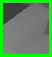}
			\includegraphics[width=1.02in]{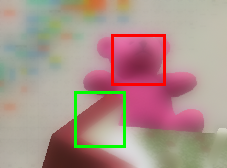}
			\includegraphics[width=0.53in]{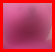}
			\includegraphics[width=0.455in]{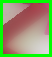}
	\end{minipage}}
	\subfigure[MS-WGF]{
		\begin{minipage}{1.02in}
			\centering
			\includegraphics[width=1.02in]{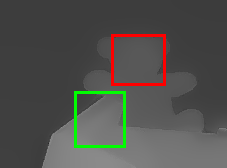}
			\includegraphics[width=0.53in]{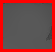}
			\includegraphics[width=0.455in]{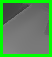}
			\includegraphics[width=1.02in]{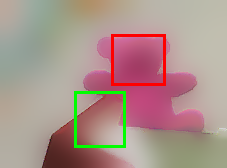}
			\includegraphics[width=0.53in]{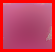}
			\includegraphics[width=0.455in]{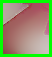}
	\end{minipage}}
	\caption{Mutual structure extraction on depth and RGB image pairs ($r=5, \epsilon=0.05, iteration=10, \lambda = 1000$).}
	\label{mutualstructuredepth}
\end{figure}

In this section, the S-WGF is applied to the iteration framework of \cite{msfjf}, which was proposed to extract structures co-existed in a reference image and a target image, to form a new filter termed Mutual-struture S-WGF (MS-WGF). The method in \cite{msfjf} is referred to as MJF. We apply MJF and MS-WGF to extract the mutual structures of an RGB image and a depth image. The results are shown in Fig.~\ref{mutualstructuredepth}. The results of MJF is obtained with 20 iterations (without post processing by median filter) and the results of MS-WGF is obtained with 10 iterations. These results demonstrate that with fewer iterations, MS-WGF performs as well as MJF. Moreover, the results on the depth image of MS-WGF is smoother than that of MJF and the non-mutual structures on the bear's face are removed more thoroughly by MS-WGF.

\subsection{Colorization}
In addition to image filtering, our new side window technique can also be used to improve other local patch based algorithms, such as colorization by optimization \cite{colorization}. The algorithm works in the $YUV$ color space. For  a given intensity channel $Y(i)$ as input, the algorithm outputs two color channels $U(i)$ and $V(i)$, where $i$ denotes a particular pixel, through optimizing the following cost function 

\begin{equation}\label{optimization}
J(U) = \mathop{\sum}_{i}\Big(U(i) - \sum_{j\in{N(i)}}\omega_{ij}U(j)\Big)^2
\end{equation}

To implement the above colorization by optimization method in the SWF framework, we simply change the neighborhood {$N(i)$ to a side window of $i$ (denoted to as $N_s(i)$) and keep all other aspects of the algorithm intact. Instead of a neighborhood centered on $i$, we choose a suitable side window $N_s(i)$ that aligns its side or corner with $i$. For each pixel, the best side window is chosen with a box filtering kernel (eq. 4). 
	
	\if false
	\begin{algorithm}[!htbp]
		\caption{Colorization based on SWF} \label{algo_color}
		\begin{algorithmic}[1]
			\REQUIRE $Y(i)$ is the intensity of pixel $i$. $S(t)$ is the set of pixels whose colors are specified: $U(t)=u_t$ and $V(t)=v_t$.
			\STATE Choose the best side window for each pixel with a box filtering kernel (equation 4). The best side window for pixel $i$ is denoted as $N_s(i)$.
			\STATE Calculate the weight of pixel $j\in N_s(i)$ 
			
			$\omega_{ij} = e^{-(Y(i)-Y(j))^2/{2\sigma^2}}$
			\STATE Solve equation (\ref{optimization}) for the $U$ and $V$ channels.
		\end{algorithmic}
	\end{algorithm}
	\fi
	Experiments have been carried out based on the image data and code provided in the web page of the authors of \cite{colorization}. Some results are shown in Fig.~\ref{colorization}. From the zoomed in patches, we can see that color leakage exists in the original method. But it is avoided when the original method is embedded with the side window technique. This is owing to the edge-preserving property of side window technique and demonstrating the wider applicability of the new side window technique. 
	
	\begin{figure}[ht!]
		\centering
		\subfigure[Input]{
			\begin{minipage}{1.02in}
				\centering
				\includegraphics[width=1.02in]{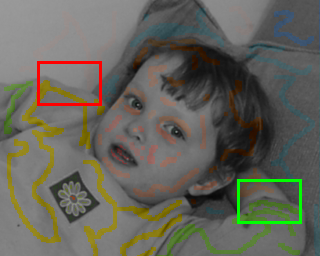}
				\includegraphics[width=1.02in]{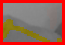}
				\includegraphics[width=1.02in]{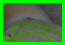}
		\end{minipage}}
		\subfigure[method in \cite{colorization}]{
			\begin{minipage}{1.02in}
				\centering
				\includegraphics[width=1.02in]{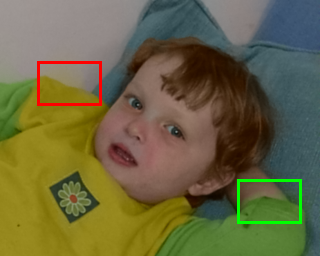}
				\begin{overpic}[width=1.02in]{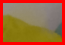}
					\put(50,50){\color{black}\vector(-2.5,-2.5){10}}
				\end{overpic}
				\begin{overpic}[width=1.02in]{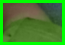}
					\put(80,50){\color{black}\vector(-2.5,-2.5){10}}
				\end{overpic}
		\end{minipage}}
		\subfigure[ours]{
			\begin{minipage}{1.02in}
				\centering
				\includegraphics[width=1.02in]{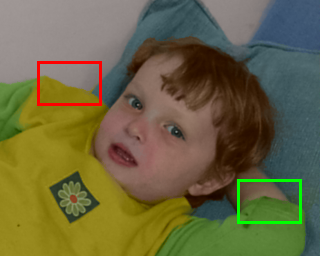}
				\begin{overpic}[width=1.02in]{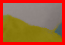}
					\put(50,50){\color{black}\vector(-2.5,-2.5){10}}
				\end{overpic}
				\begin{overpic}[width=1.02in]{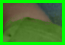}
					\put(80,50){\color{black}\vector(-2.5,-2.5){10}}
				\end{overpic}
		\end{minipage}}
		\subfigure[Input]{
			\begin{minipage}{1.02in}
				\centering
				\includegraphics[width=1.02in]{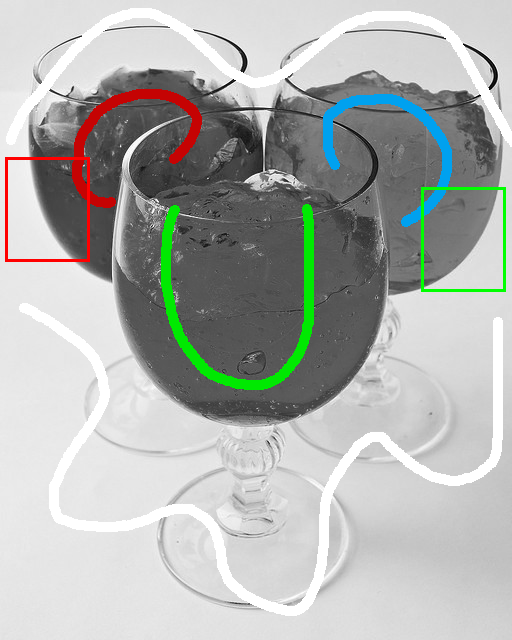}
				\includegraphics[width=0.49in]{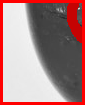}
				\includegraphics[width=0.49in]{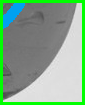}
		\end{minipage}}
		\subfigure[method in \cite{colorization}]{
			\begin{minipage}{1.02in}
				\centering
				\includegraphics[width=1.02in]{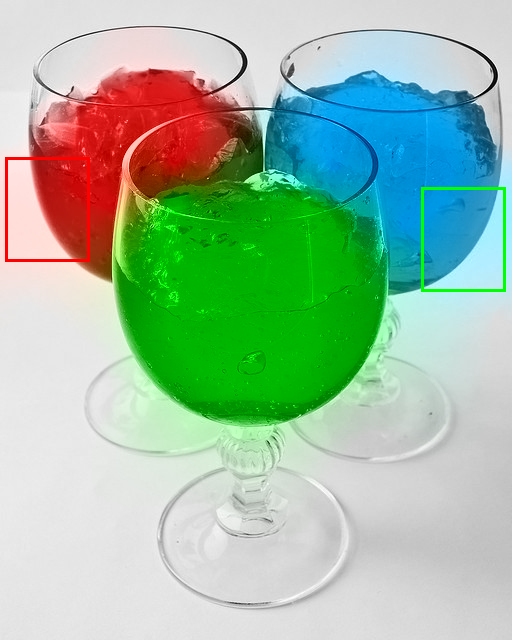}
				\begin{overpic}[width=0.49in]{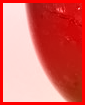}
					\put(35,35){\color{black}\vector(-2.5,-2.5){15}}
				\end{overpic}
				\begin{overpic}[width=0.49in]{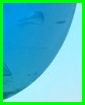}
					\put(50,35){\color{black}\vector(2.5,-2.5){15}}
				\end{overpic}
		\end{minipage}}
		\subfigure[ours]{
			\begin{minipage}{1.02in}
				\centering
				\includegraphics[width=1.02in]{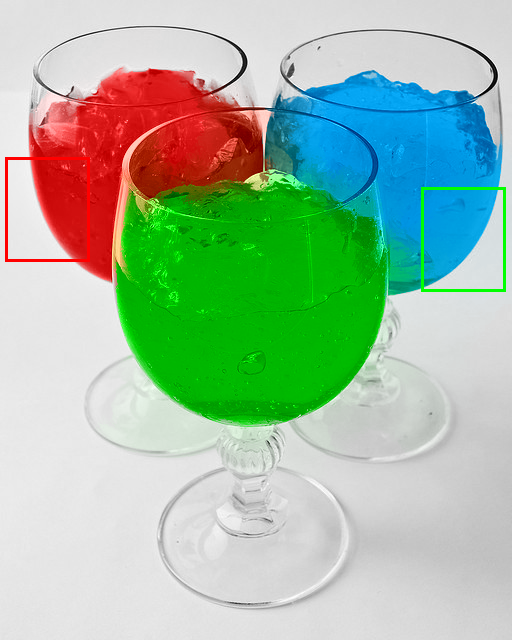}
				\begin{overpic}[width=0.49in]{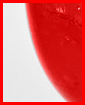}
					\put(35,35){\color{black}\vector(-2.5,-2.5){15}}
				\end{overpic}
				\begin{overpic}[width=0.49in]{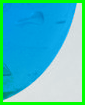}
					\put(50,35){\color{black}\vector(2.5,-2.5){15}}
				\end{overpic}
		\end{minipage}}
		\caption{Colorization ($r=3$). Color leakage existed in the original method is avoided by implementing the method under the SWF framework.}
		\label{colorization}
	\end{figure}
	
	\section{Complexity Analysis}
	The order of complexity of side window based filters is the same as the original filters. However, as the SWF implementation needs to perform calculations over multiple windows, its computational cost is higher. 
	Our experiments without code optimization are conducted on a computer with a 3.5GHz Intel core Xeon(R) CPU. For gray-scale images with 1 mega pixels, the computational time of the filters are shown in Table ~\ref{computationtime}. The BIL is the original algorithm in \cite{bf} without modification for acceleration. With code optimization and implementing GPU programming, the computational speed can be significantly improved. Our code will be made available publicly after the paper is published.
	\begin{table}
		\centering
		\caption{The computational time on images with 1 mega pixels}
		\begin{tabular}{|c|c|c|c|c|c|}
			\hline
			Method&BOX&GAU&MED&BIL&GUI\\
			\hline
			Original&0.052&0.023&1.16&8.69&0.131\\
			\hline
			SWF version&0.215&0.23&3.67&26.2&0.431\\
			\hline
		\end{tabular}
		\label{computationtime}
	\end{table}
	\section{Conclusion}
	Window based processing is one of the most common operations in computer vision. Traditional practices almost always align the center of the window with the pixel under processing. In this paper, we show that this widely used practice is not always the best solution. We show that in many applications, the side or the corner of the operation window instead of the center should be aligned with the pixel under processing and propose the side window filtering (SWF) technique. We have shown that many popular linear and non-linear filtering algorithms can be implemented based on this principle and the SWF implementation of these traditional filters can significantly boost their edge preserving capabilities. We have further shown that the SWF principle can be extended to other computer vision problems that involve a local operation window and a linear combination of the neighbors in this window such as colorization by optimization. We have shown that SWF technique can improve their performances and avoid artifacts such as color leakage that is often associated with such algorithm. Window based operations is extensively used in many areas of computer vision and machine learning including convolutional neural networks (CNNs). The SWF principle, i.e., aligning the edge or corner of the operation window with the pixel being processed, although seemingly trivial, is actually deeply rooted in the fundamental assumptions of many algorithms. Our theoretical analysis and state of the art results for many real world applications have demonstrated its effectiveness. We believe that there are many more applications can benefit from implementing the SWF principle.    

{\small
\bibliographystyle{ieee_fullname}
\bibliography{egpaper_final}
}

\end{document}